%% file: main.tex
\documentclass{fairmeta}
\usepackage{fancyhdr}
\usepackage{amsmath}
\usepackage{enumerate} 
\usepackage{bm}
\usepackage{algorithm}
\usepackage{floatflt}
\usepackage{algpseudocode}
\usepackage{amsfonts}
\usepackage{amsthm}
\usepackage{newtxtt}
\usepackage{algorithm}
\usepackage{colortbl} 
\usepackage{cleveref}
\usepackage{diagbox}
\usepackage[utf8]{inputenc}
\usepackage{textgreek}
\usepackage{colortbl}
\usepackage{nicematrix}
\usepackage{makecell}
\usepackage{float}
\usepackage{arydshln}
\usepackage[frozencache,cachedir=.]{minted}
\usepackage[justification=centering]{caption}
\usepackage{subcaption}
\usepackage{pgfplots}
\pgfplotsset{compat=1.17}
\usepackage{amssymb}
\usepackage{xspace}
\usepackage{wrapfig}
\usepackage{tabularx}
\usepackage{booktabs}
\usepackage{mathtools}
\usepackage{wrapfig}
\usepackage{amssymb}

\usepackage{array}
\usepackage{xcolor}
\usepackage{pifont}
\usepackage{graphicx}

\usepackage{enumitem}
\usepackage[export, valign=t]{adjustbox}

\usepackage[T1]{fontenc}
\usepackage{lmodern}

\usepackage{listings}

\usepackage[most]{tcolorbox}

\lstdefinelanguage{YAML}{
    keywords={true,false,null},
    keywordstyle=\color{black},
    sensitive=false,
    comment=[l]{\#},
    commentstyle=\color{gray},
    stringstyle=\color{black},
    morestring=[b]",
    morestring=[b]'
}

\lstdefinestyle{mystyle}{
    backgroundcolor=\color{lightgray},
    basicstyle=\scriptsize\ttfamily,
    breakatwhitespace=false,
    breaklines=true,
    captionpos=b,
    commentstyle=\color{black},
    keywordstyle=\color{black},
    language=YAML,
    numbers=left,
    numbersep=5pt,
    numberstyle=\tiny\color{gray},
    showstringspaces=false,
    stringstyle=\color{black},
    tabsize=4,
}

\newtcblisting{mycodebox}[2][]{%
    colback=lightgray,
    colframe=black,
    listing only,
    listing options={style=mystyle},
    breakable,
    title=#2,
    #1
}

\usepackage{silence}
\usepackage{ulem}
\makeatletter
\patchcmd{\wrong@fontshape}{\@gobbletwo}{}{}{}
\makeatother
\usepackage{mwe} 
\newtheorem{theorem}{Theorem}[]

\newtheorem{remark1}[theorem]{Remark}

\definecolor{upColor}{RGB}{17,138,21}
\definecolor{downColor}{RGB}{174,36,67}

\makeatletter
\def\smartfiguretype{figure}
\newcommand{\smartmention}[2]{%
  \ifx\@captype\smartfiguretype
    \textbf{#2}%
  \else
    \ifcsname smartseen#1\endcsname
      \textit{#2}%
    \else
      \textbf{#2}%
      \expandafter\gdef\csname smartseen#1\endcsname{}%
    \fi
  \fi
  \xspace
}
\makeatother

\newcommand{\ours}{\smartmention{smart}{SMART}}
\newcommand{\ourwork}{\smartmention{smart}{SMART}}
\newcommand{\ourframework}{\smartmention{smart-sim}{SMART-Sim}}
\newcommand{\ourdataset}{\smartmention{smart-data}{SMART-Data}}

\newcommand{\THEPT}{\smartmention{smart-vla}{SMART-VLA}}
\newcommand{\thept}{\smartmention{smart-vla}{SMART-VLA}}
\newcommand{\THEFS}{\smartmention{smart-vla-no-s1}{SMART-VLA(w/o S1)}}
\newcommand{\thefs}{\smartmention{smart-vla-no-s1}{SMART-VLA(w/o S1)}}
\newcommand{\THEBASE}{\smartmention{smart-vla-real}{SMART-VLA(Real)}}
\newcommand{\thebase}{\smartmention{smart-vla-real}{SMART-VLA(Real)}}

\newcommand{\thebaseline}{$\pi_{0.5}$\xspace}

\definecolor{molmocol}{HTML}{FBEAF2}
\definecolor{ourblue}{RGB}{220,235,250}
\definecolor{checkcol}{HTML}{45BFC0}
\definecolor{crosscol}{HTML}{FF7DB5}

\newcommand{\cmark}{\textcolor{green!60!gray!80!white!80}{\ding{51}}}
\newcommand{\xmark}{\textcolor{purple!30!pink!90}{\ding{55}}}

\fancypagestyle{gammaTeleAI}{%
  \fancyhf{}
  \fancyhead[C]{\small\sffamily Gamma Robotics / TeleAI}
  \fancyfoot[C]{\thepage}
  \renewcommand{\headrulewidth}{\ifnum\value{page}>1 0.3pt\else 0pt\fi}
  
}
\makeatletter
\let\ps@plain\ps@gammaTeleAI
\makeatother

\title{
SMART: Zero-Shot Sim-to-Real Articulated Object Manipulation via Large-Scale Synthetic Pretraining
}
\author[1\dagger\ddagger]{\text{Jicong Ao}}
\author[1,2\ddagger]{\text{Shuhan Jiang}}
\author[1,3]{\text{Yuling Zhong}}
\author[1,3]{\text{Yanwen Liu}}
\author[1,4]{\text{Yuhan Gao}}
\author[1,5]{\text{Jiangyuan Zhao}}
\author[1,6]{\text{Yang Zhang}}
\author[2]{\text{Shiqiang Zhu}}
\author[1,7*]{\text{Chenjia Bai}}
\author[1*]{\text{Xuelong Li}}
\affiliation[1]{Institute of Artificial Intelligence, China Telecom}
\affiliation[2]{Zhejiang University}
\affiliation[3]{Technical University of Munich}
\affiliation[4]{Harbin Institute of Technology}
\affiliation[5]{Shanghai Jiao Tong University}
\affiliation[6]{Tsinghua University}
\affiliation[7]{Gamma Robotics ($\gamma$)}
\contribution{\textsuperscript{$\dagger$}Project Lead$\quad$\textsuperscript{$\ddagger$}Equal Contributions$\quad$\textsuperscript{*}Corresponding Authors
}
\date{October, 2026}
\project{\url{https://teamillusion-smart.github.io}}
\dataset{\url{https://huggingface.co/datasets/TeleEmbodied/SMART-Data}}
\metadata[Correspondence to]{Chenjia Bai (\email{baicj@chinatelecom.cn})}

\input{sections/00_abstract}

\begin{document}

\maketitle

\input{sections/01_introduction}

\input{sections/02_related_works}

\input{sections/03_framework}

\input{sections/04_data_synthesis}

\input{sections/05_simulation_experiments}
\input{sections/06_real_world_experiment}

\input{sections/07_discussion}

\input{sections/08_contributions}

\bibliographystyle{unsrtnat}
\bibliography{references,auto_reference} 

\input{sections/99_appendix}

\end{document}

%% file: sections/00_abstract.tex
\abstract{
The ability to interact with articulated objects is essential for embodied intelligent systems, but collecting large-scale real-world demonstrations for these interactions remains challenging due to the precise contact and constraint-following motions involved.
Although simulation provides a promising alternative, existing synthetic data efforts cover limited articulated-object categories, while general-purpose synthesis pipelines lack explicit designs for part-level semantics and articulation constraints, hindering agentic task generation and scalable synthesis of high-quality articulated-manipulation demonstrations.
To bridge this gap, we introduce \ourwork, a scalable system leveraging
large-scale \textbf{S}ynthesized \textbf{M}anipulation demonstrations for
\textbf{ART}iculated-object manipulation.
At its core, we develop \ourframework, a simulation platform with articulation-aware design that enables effective task generation and efficient demonstration collection.
Building on \ourframework, we apply agentic task generation and design a scalable distributed synthesis system, using them to synthesize \ourdataset, comprising over 1M demonstrations across 44 atomic task types, 5 robot setups, and 2{,}507 articulated objects.
The vision-language-action (VLA) model pretrained on \ourdataset shows competitive performance on simulation benchmarks
and achieves zero-shot sim-to-real transfer and scalable performance in real-world articulated-object manipulation tasks.
This highlights the potential of synthetic demonstrations in providing effective and scalable supervision for improving VLA model performance in contact-rich articulated-object manipulation.
}

%% file: sections/01_introduction.tex


\section{Introduction}

\begingroup
\renewcommand{\smartmention}[2]{\textbf{#2}\xspace}

\begin{figure}[tp]
    \centering
    \includegraphics[width=1.0\linewidth]{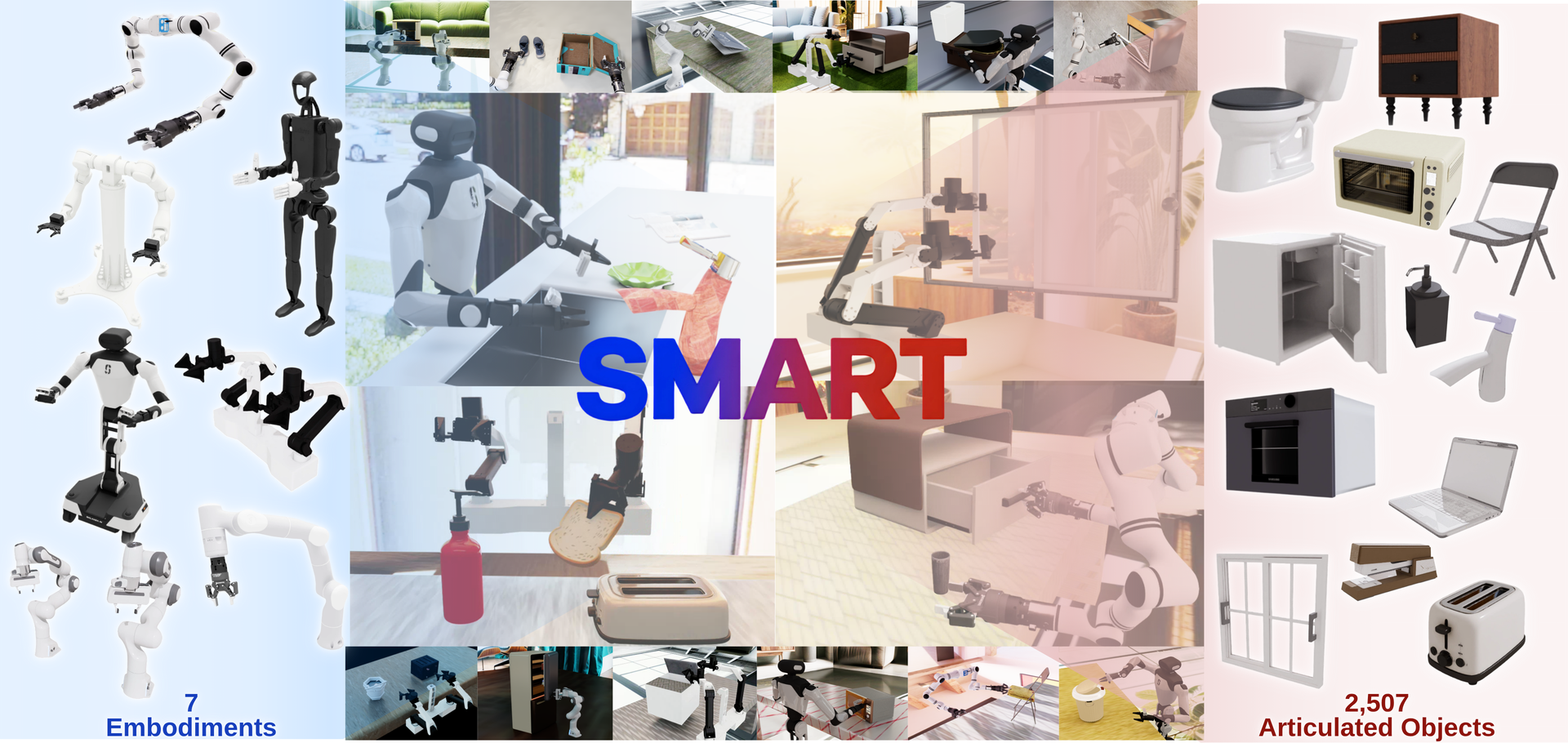}
    \caption{
    \ours synthesizes large-scale simulation data for articulated-object manipulation by combining diverse embodiments and articulated objects, rich manipulation skills, and comprehensive domain randomization.
    }
    \label{fig:teaser}
\end{figure}

Articulated objects are ubiquitous in human environments and central to many everyday activities.
Their movable parts enable rich functionality while imposing structured geometric and physical constraints on interactions.
Compared with rigid-object manipulation, articulated-object manipulation requires reasoning about functional parts and joint-constrained motions while maintaining precise and sustained contact throughout the interaction~\citep{ma2023sim2real2,wu2025artibench,wang2025articubot}.
Enabling robots to engage in such interactions can facilitate the emergence of intelligent behaviors, making it an important step toward embodied intelligence.

Traditional approaches to articulated-object manipulation often rely on explicit object models, online kinematic estimation, and affordance prediction~\citep{ma2023sim2real2,jiang2025dexsim2real2,wang2024onlineaxis,wu2023environmentaware,ling2024coarsetofine,chen2026pa3ff}, which require task-specific assumptions and limit scalability across diverse objects and long-horizon interactions.
Recently, vision-language-action (VLA) models have emerged as a promising paradigm for general-purpose manipulation~\citep{brohan2023rt1,zitkovich2023rt2,kim2025openvla,black2025pi0}, with their capability largely shaped by the scale, diversity, and quality of training demonstrations~\citep{black2025pi0,nvidia2025grootn1,wu2026lingbotvla}.
Yet such data is still predominantly collected through human teleoperation~\citep{wu2025robocoin,bu2025agibot}, which is costly and hard to scale.
To alleviate the burden of robot-based data collection, Universal Manipulation Interface~\citep{chi2024umi,zhaxizhuoma2025fastumi} and egocentric approaches~\citep{hoque2026egodex,kareer2025egomimic} have been proposed to collect demonstrations through handheld devices or human-centric embodiments.
However, transferring these demonstrations to robot platforms introduces embodiment mismatches and physical infeasibilities~\citep{zhang2026hyembodied05vla,yang2025egovla,zheng2026egoscale}, which can degrade demonstration quality.
These challenges are particularly pronounced in articulated-object manipulation, where effective demonstrations require precise and sustained contact with functional parts while satisfying articulation constraints.
As a result, high-quality real-world demonstrations for articulated-object manipulation remain difficult to collect at scale.

Simulation offers a natural alternative to real-world demonstration collection.
Prior works~\citep{mandlekar2023mimicgen, nasiriany2026robocasa365, li2023behavior1k} show that simulation can scale demonstration collection at low cost.
Recent systems such as InternData-A1~\citep{tian2025interndataa1} and MolmoBot~\citep{deshpande2026molmobot} further highlight this potential by generating large-scale synthetic datasets for general manipulation and evaluating sim-to-real transfer using only simulated data.
Despite this progress, existing simulation-based datasets for articulated-object manipulation often cover only a narrow range of object categories or remain limited in scale~\citep{cui2025gapartmanip,wang2025articubot}.
Although works such as InternData-A1~\citep{tian2025interndataa1} and MolmoBot~\citep{deshpande2026molmobot} demonstrate sim-to-real transfer across several manipulation tasks, their synthetic data covers only a limited range of articulated-object categories.
In addition, their general-purpose demonstration synthesis pipelines do not explicitly account for part-level semantics and articulation constraints, leading to limited adoption of agentic task generation and challenges in synthesizing high-quality demonstrations for articulated-object manipulation.
Consequently, large-scale synthetic data for VLA learning and sim-to-real transfer across diverse articulated-object manipulation remains underexplored.

To fill this gap, we propose \ours,
a scalable system leveraging \textbf{S}ynthetic \textbf{M}anipulation data for \textbf{ART}iculated object manipulation.
We first introduce \ourframework,
a simulation platform with specialized articulation-aware designs for rapid task specification and effective motion generation.
We further employ agentic task generation to create articulated-object manipulation tasks with varying complexity and temporal horizons. Combined with our distributed data synthesis system, this enables efficient and scalable generation of articulated manipulation tasks and trajectories, resulting in \ourdataset, a large-scale synthetic dataset covering diverse articulated objects and manipulation skills.
We conduct behavior cloning pretraining exclusively on \ourdataset, with the pretraining effect first evaluated on simulation benchmarks.
To further explore \ourframework's potential in data synthesis, we conduct zero-shot sim-to-real experiments and performance scaling experiments on representative articulated-object manipulation tasks.
Our contributions are summarized as follows:

\begin{itemize}[leftmargin=*,
nosep
]

    \item We propose \ourframework, a scalable simulation platform with specialized articulation-aware designs for asset annotation, motion generation, skill design, and domain randomization, enabling efficient and effective task and motion generation for articulated-object manipulation.

    \item We develop an \textbf{agentic task generation approach} and a \textbf{distributed synthesis system}, and use them to construct \ourdataset, a large-scale synthetic articulated-object manipulation dataset with over 1M trajectories across 5 robot setups and 23 object categories, providing diverse supervision for articulated-object manipulation learning.

    \item Extensive simulation and real-world experiments show that pretraining on \ourdataset can improve VLA performance on downstream tasks.
    Furthermore, data synthesized by \ourframework enables zero-shot sim-to-real policy transfer with positive scaling trends, highlighting the potential of synthetic data as a scalable supervision source for contact-rich articulated-object manipulation.

\end{itemize}
\endgroup

%% file: sections/02_related_works.tex
\section{Related Work}

\textbf{Articulated object manipulation.}
Compared to rigid-object manipulation, articulated-object manipulation requires additional reasoning about functional parts, joint constraints, and sustained contact during interaction. 
Early works address these challenges by leveraging pretrained encoders to extract auxiliary information such as the articulation axis~\citep{wang2024onlineaxis} and affordance~\citep{wu2023environmentaware,ling2024coarsetofine} for manipulation.
Recognizing the value of simulation in providing accurate articulation models and supervision, GAPartManip~\citep{cui2025gapartmanip} and Infinigen-Articulated~\citep{joshi2025infinigenarticulated} develop curated articulated-object datasets and automated pipelines for generating simulation-ready assets with articulation annotations.
Building on such simulated articulated-object models, subsequent works either construct explicit models for manipulation~\citep{ma2023sim2real2, jiang2025dexsim2real2} or train more capable affordance prediction models~\citep{chen2026pa3ff}.
Despite their effectiveness, these approaches often rely on simplified object models and accurate perception of object geometry and articulation, limiting their scalability to diverse objects and real-world settings.

\textbf{VLA models and Real-world data.}
Vision-language-action models have recently become a prominent approach to general-purpose robot control, and their performance depends on high-quality, diverse, and large-scale demonstration datasets.
Works like the $\pi$ series~\citep{black2025pi0, physicalintelligence2025pi05}, LingBot-VLA~\citep{wu2026lingbotvla}, and PRTS~\citep{zhang2026prts} heavily rely on teleoperated robot demonstrations, using more than 10,000 hours of multi-robot data, approximately 20{,}000 hours of real bimanual data, and 167 billion tokens of manipulation and embodied-reasoning data, respectively.
However, teleoperation remains hard to scale because of its high cost and low data collection efficiency.
To reduce this burden, studies on UMI~\citep{chi2024umi,zhaxizhuoma2025fastumi} enable portable robot-free demonstration collection through handheld interfaces, with Hy-Embodied-0.5-VLA~\citep{zhang2026hyembodied05vla} further performing effective pretraining with more than 10{,}000 hours of UMI data.
Additionally, egocentric demonstration collection offers another robot-free alternative by capturing human wrist and hand motion, with Ego4D~\citep{grauman2022ego4d} and EgoDex~\citep{hoque2026egodex} providing egocentric human interaction data and EgoMimic~\citep{kareer2025egomimic}, EgoVLA~\citep{yang2025egovla}, and EgoScale~\citep{zheng2026egoscale} demonstrating its value for VLA pretraining.
Despite improved collection efficiency and lower cost, these methods either face a trade-off between tracking precision and portability or require pose estimation, retargeting, and robot alignment to recover executable actions.
These limitations are particularly problematic for articulated-object manipulation, where small action errors can break sustained contact or violate joint constraints.
Overall, current real-world demonstration collection methods struggle to simultaneously provide high-quality data, broad diversity, and high collection efficiency for articulated-object manipulation tasks.

\textbf{Simulation and synthesized data for robot manipulation.}
Simulation offers a controllable and repeatable setting for generating executable robot trajectories.
Frameworks such as Isaac Lab~\citep{nvidia2025isaaclab} and robosuite~\citep{zhu2020robosuite} provide reusable assets and frameworks for robot-learning experiments.
Building on these infrastructures, SceneFoundry~\citep{chen2026scenefoundry} and SceneSmith~\citep{pfaff2026scenesmith} provide methods to construct interactive, simulation-ready scenes for simulation tasks, while RoboCasa~\citep{nasiriany2024robocasa}, RoboCasa365~\citep{nasiriany2026robocasa365}, and BEHAVIOR-1K~\citep{li2023behavior1k} provide household environments and task suites with broad scene and task coverage.
MimicGen~\citep{mandlekar2023mimicgen}, RoboTwin 2.0~\citep{chen2025robotwin2}, and HumanoidGen~\citep{jing2025humanoidgen} then scale trajectory generation through demonstration recombination, automatic task programming, and constraints generated by a large language model (LLM) with trajectory optimization, respectively.
At the dataset level, InternData-A1~\citep{tian2025interndataa1} and MolmoBot~\citep{deshpande2026molmobot} use 7,433 and 5,704 hours of synthetic data for VLA pretraining,
showing that large synthetic datasets can support policy pretraining and real-world transfer.
However, large-scale synthetic data for VLA learning remains underexplored for diverse articulated-object manipulation, especially for contact-intensive tasks that demand precise contact and constrained motion.

%% file: sections/03_framework.tex
\section{Simulation Platform}
\label{sec:simulation_platform}

\begin{figure}[t!]
    \centering
    \includegraphics[width=1\linewidth]{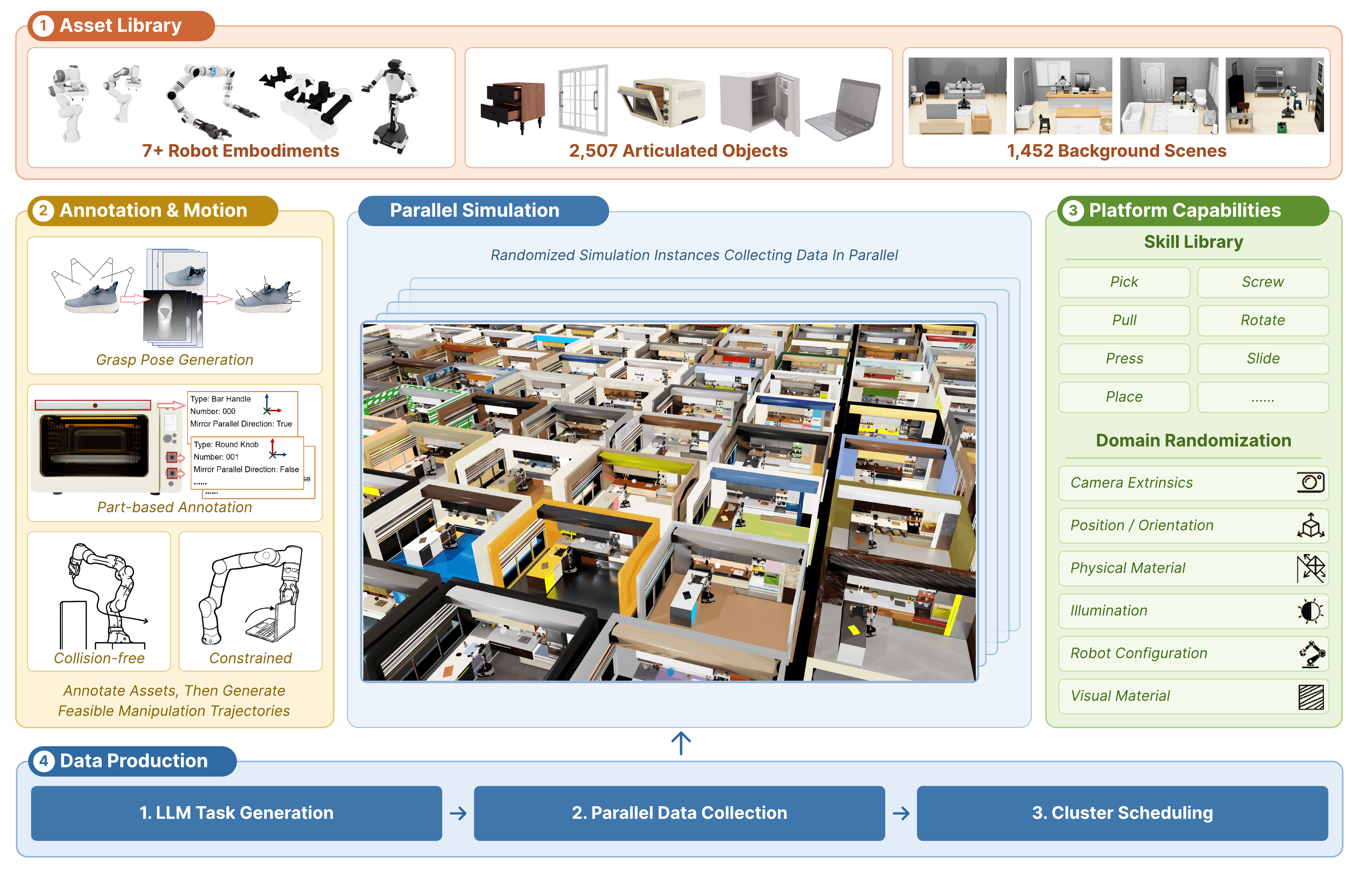}
    \caption{
    Overview of \ourframework. 
    \ourframework provides versatile robot embodiment and asset support, flexible annotation and skills for motion generation, efficient demonstration collection, and comprehensive domain randomization features.
    }
    \label{fig:framework}
\end{figure}

To enable efficient robot manipulation data collection, we propose \textbf{SMART-Sim}, a simulation platform built on Isaac Lab~\citep{nvidia2025isaaclab} that streamlines demonstration synthesis across diverse manipulation tasks.
As illustrated in Fig.~\ref{fig:framework}, \ourframework provides versatile support for robot embodiments and assets, flexible annotation and motion generation, efficient demonstration collection, and comprehensive domain randomization.
We further provide a comprehensive comparison with existing simulation platforms in Tab.~\ref{tab:sim-framework-comparison}.

Notably, \ourframework introduces articulation-aware designs throughout asset annotation, motion generation, skill definition, and domain randomization.
These designs account for the multi-part structure, semantic roles, and physical constraints of articulated objects, enabling efficient collection of high-quality demonstrations.

\input{tables/framework_comparison}

In this comparison, the Scenes row counts distinct simulation environments, and the Embodiments row counts supported robot platforms, while rigid and articulated objects are reported separately when such a distinction is available.
Benefiting from the annotation and motion-generation pipeline, \ourframework provides the largest articulated-object asset base among the compared frameworks and is the only one that combines AI-generated tasks, multi-environment parallelism, and distributed data synthesis in a single pipeline.

\subsection{Simulation Assets}
\label{subsec:simulation_assets}

\textbf{Embodiments.}
\ourframework currently supports \textbf{seven} robot embodiments: Franka Emika Panda, RM75, ARX AC-1 (AC1), R1Pro, Flexiv Rizon 4S, Marvin M6s, and Unitree H1-2.
It also provides a setup assembly module that combines robot embodiments with configurable camera layouts, allowing new hardware configurations to be created quickly.
Using this module, we construct five representative setups for data synthesis: a dual-arm RM75 setup, an R1Pro setup, an AC1 setup, and both single-arm and dual-arm Franka Panda setups.
See Tab.~\ref{tab:app:robot_platform_details} in the appendix for detailed specifications of these embodiments.

\textbf{Rigid objects.}
We select open-source assets from works such as RoboTwin~\citep{chen2025robotwin2} and GenManip~\citep{gao2025genmanip} for general-purpose manipulation, providing \textbf{9{,}116} objects across \textbf{1{,}071} categories.
The rigid-object collection also includes assets produced through 3D generation~\citep{yang2023sam3d}.
We further generate task-specific rigid objects using Hunyuan3D~\citep{zhao2025hunyuan3d} for downstream experiments.

\textbf{Articulated objects.}
We collect assets from PartNet-Mobility~\citep{xiang2020sapien} and GRU-Scene~\citep{wang2024grutopia} and process their physical properties to enable their use in \ourframework.
We also incorporate a commercially acquired, high-quality articulated object dataset to broaden the category coverage.
Our articulated object library comprises \textbf{2{,}507} objects across \textbf{88} functional categories, encompassing diverse kinematic structures and joint mechanisms for robot learning.

\textbf{Visual scenes and backgrounds.}
We use task scenes in mesh format from SDGScenes~\citep{gao2026sdgscenes}, which consists of \textbf{300} room scenes generated from diverse scene language descriptions.
To further enhance visual realism, we incorporate scenes based on 3D Gaussian Splatting (3DGS) from InteriorGS~\citep{InteriorGS2025} and GN0~\citep{li2026gn0}, yielding a total of \textbf{1{,}452} scenes.
Additionally, we generate over \textbf{10{,}000} high-resolution (\textbf{4K}) indoor images using Qwen-Image~\citep{wu2025qwenimage} as both pseudo-backgrounds and illumination sources, which are used to reduce simulation overhead and enrich lighting diversity during rendering.

\subsection{Manipulation Motion Generation}
\label{subsec:motion_gen}

\begin{figure*}[t]
    \centering

    \begin{subfigure}[t]{0.21\textwidth}
        \centering
        \includegraphics[width=\linewidth]{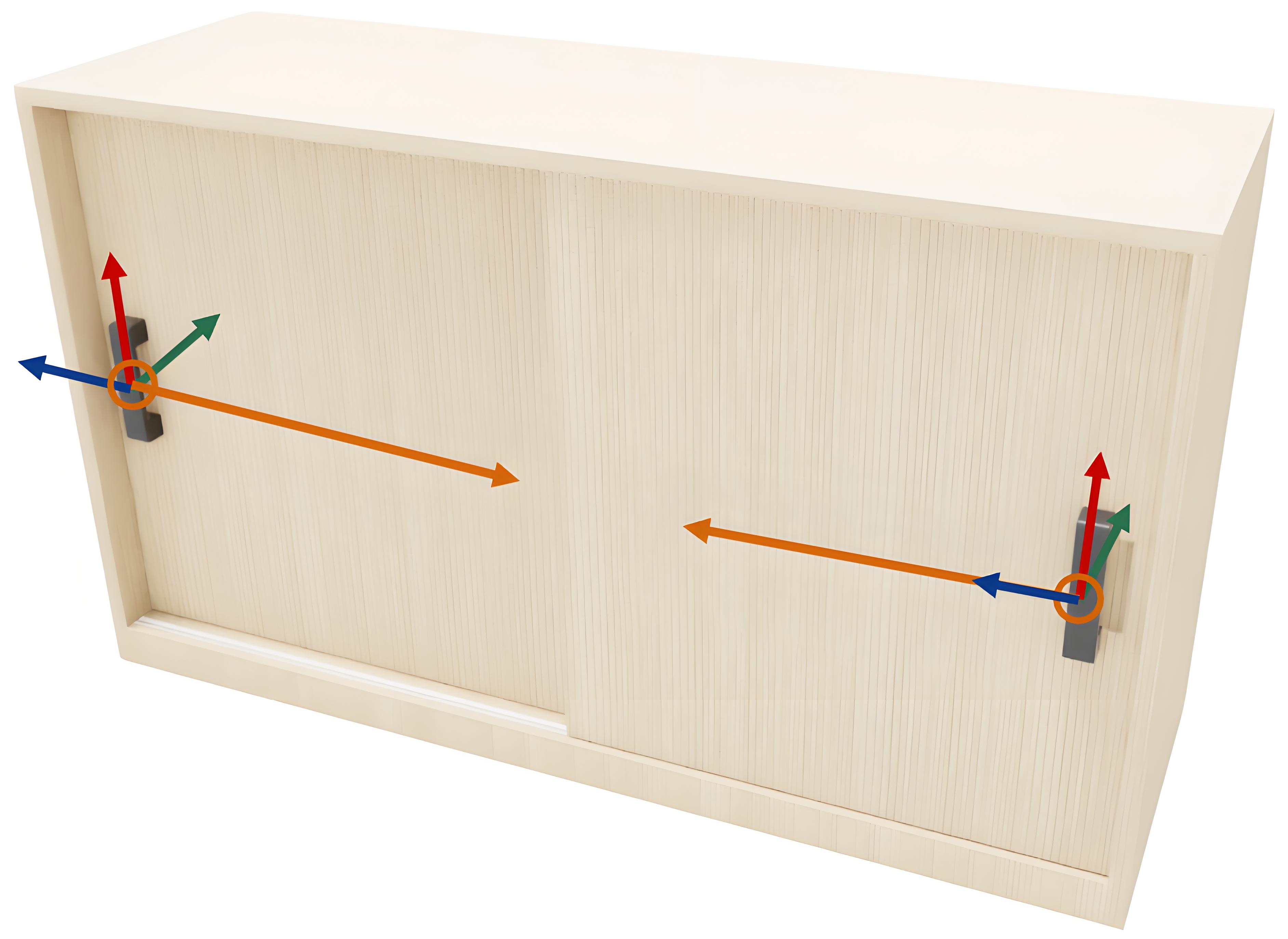}
        \caption{}
        \label{fig:sub1}
    \end{subfigure}
    \hfill
    \begin{subfigure}[t]{0.29\textwidth}
        \centering
        \includegraphics[width=\linewidth]{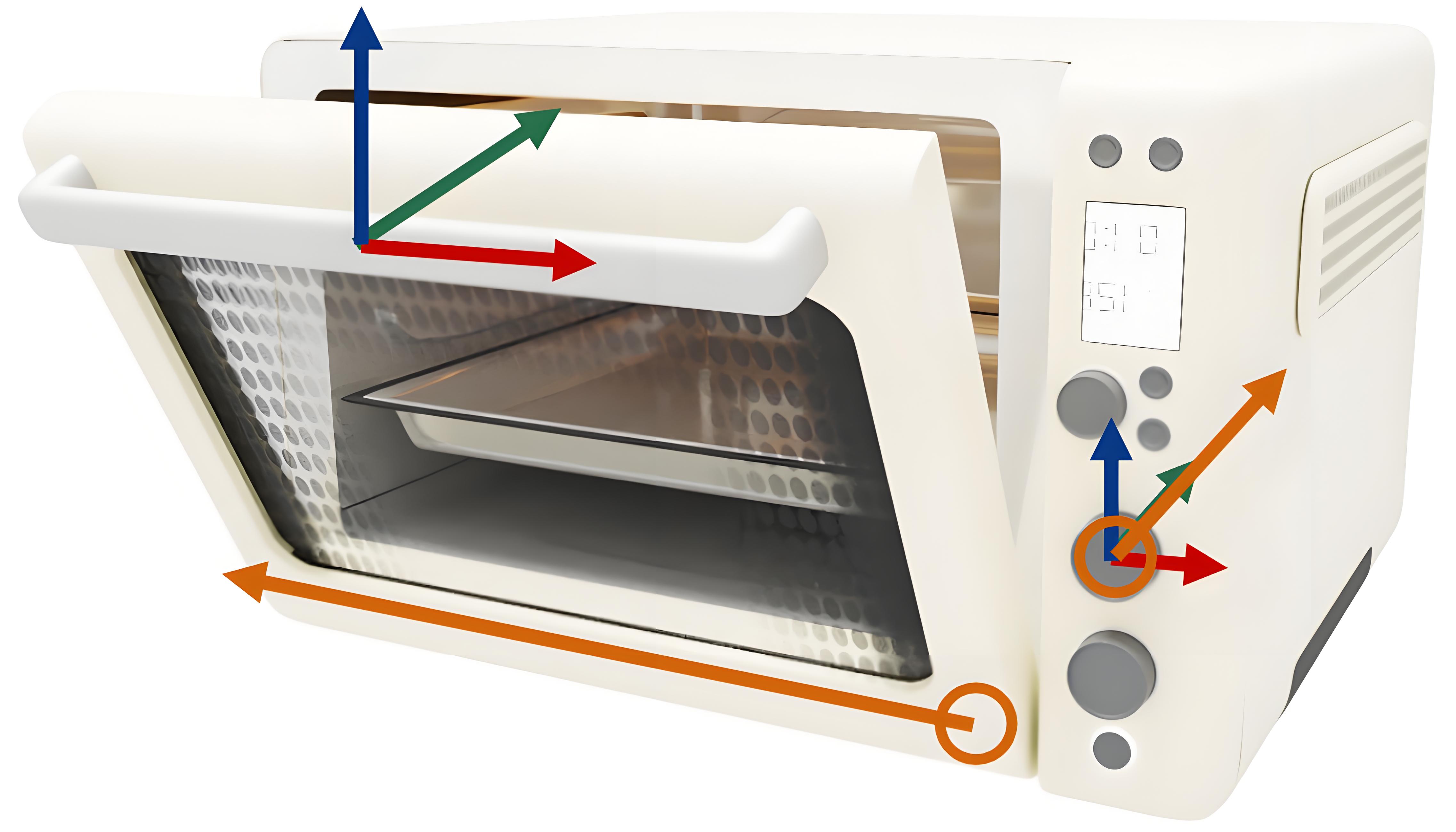}
        \caption{}
        \label{fig:sub2}
    \end{subfigure}
    \hfill
    \begin{subfigure}[t]{0.18\textwidth}
        \centering
        \includegraphics[width=\linewidth]{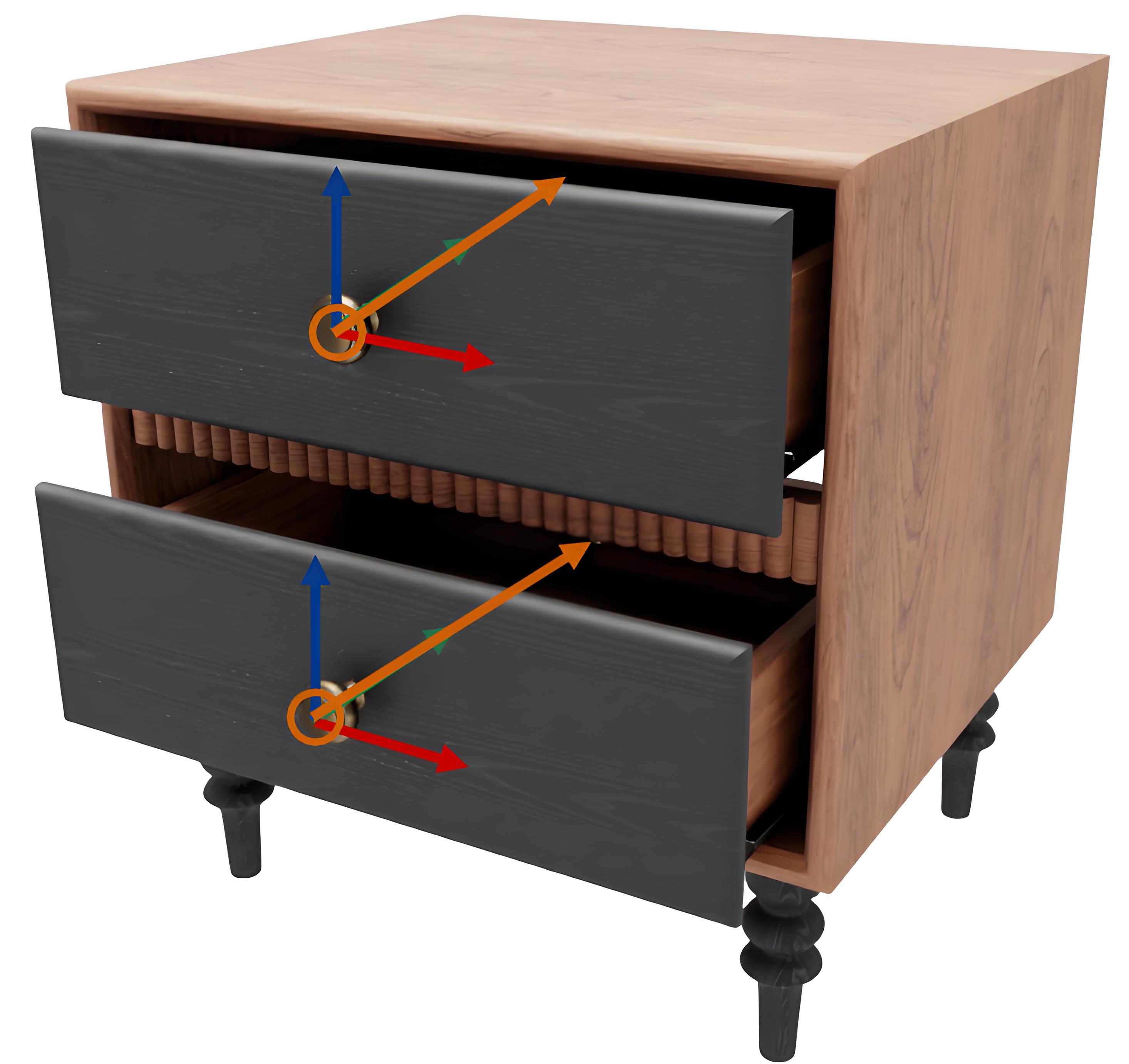}
        \caption{}
        \label{fig:sub3}
    \end{subfigure}
    \hfill
    \begin{subfigure}[t]{0.22\textwidth}
        \centering
        \includegraphics[width=\linewidth]{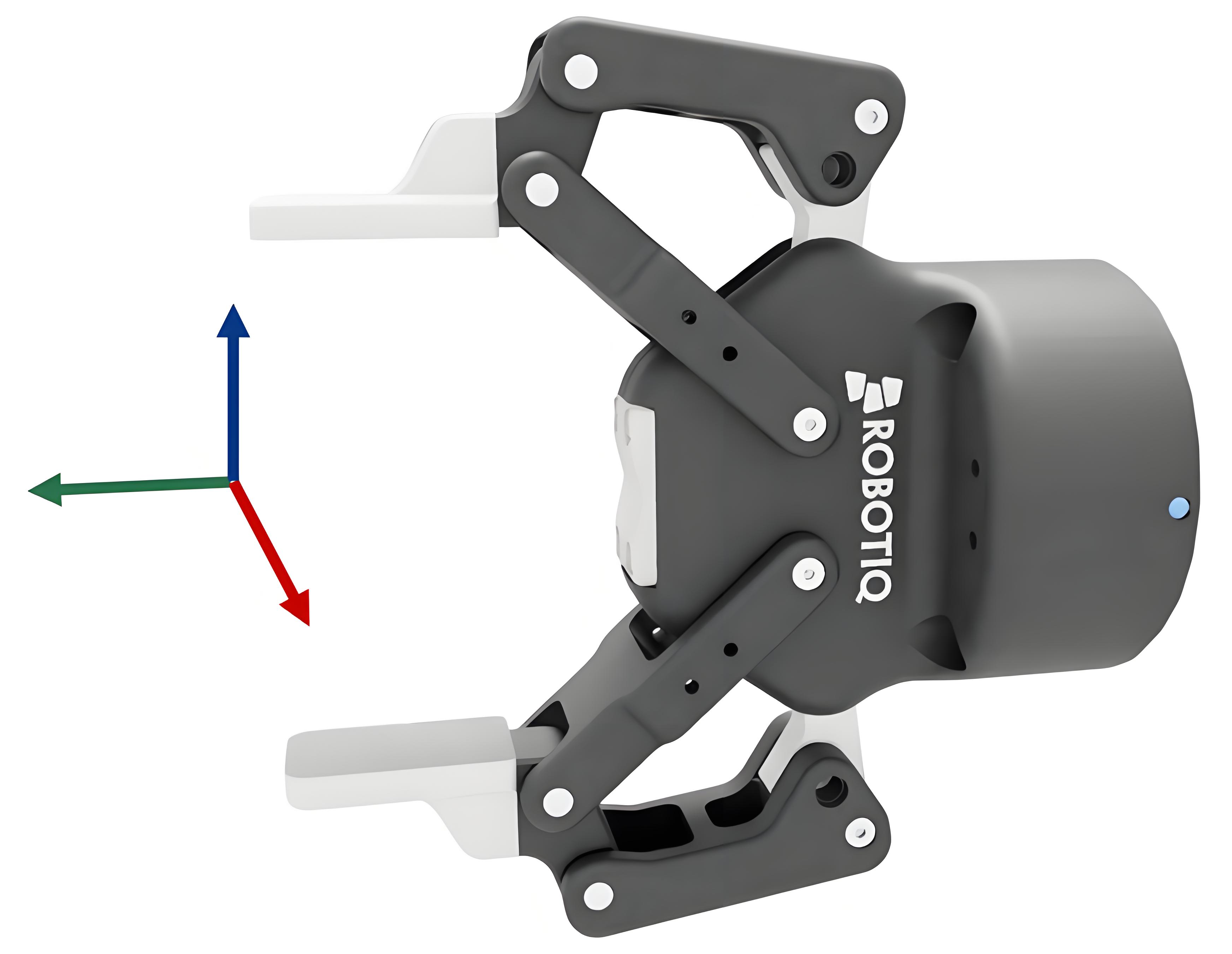}
        \caption{}
        \label{fig:sub4}
    \end{subfigure}

    \caption{The illustration of asset annotations, with (a), (b), and (c) showing the action frames on different articulated objects, and (d) showing the annotation on a Robotiq end effector. The orange arrows show the direction of the articulation constraints between parts.}
    \label{fig:annotation}
\end{figure*}

\textbf{Manipulation annotation.}
\label{subsec:manipulation_annotation}
We adopt the action frame definition in HumanoidGen~\citep{jing2025humanoidgen} and develop annotation pipelines for both rigid bodies and articulated objects to reduce manual annotation effort and accelerate asset processing.
On robot end-effectors, we define the +Z axis as the approach direction and the +Y axis as the parallel axis, leaving the +X axis fixed by the right-hand rule.
For articulated assets, the parallel axis is defined according to the joint type: it is aligned with the rotation axis for revolute joints and lies in the plane perpendicular to the translation direction for prismatic joints.
The origin of the frame is the contact point between grippers and objects, and aligning the annotations on them provides the pose targets for motion generation. 
We refer readers to HumanoidGen~\citep{jing2025humanoidgen} for a detailed description of the action-frame annotations.

For rigid objects, we use AnyGrasp~\citep{fang2023anygrasp} to generate grasp annotations.
Specifically, each object is rendered from four different viewpoints, and AnyGrasp is applied to predict 40 grasp poses for each object. 
The predicted grasp poses are then projected back to the object base coordinate frame using the extrinsic parameters of the rendering cameras. 
See Fig.~\ref{app:subfig:rigid_anno} for a detailed illustration of the approach.
In addition, object categories and language descriptions are generated by a vision-language model (VLM), which are later used for LLM-based task generation.

For articulated objects, we construct a separate pipeline to accelerate the annotation of manipulable parts. 
A VLM first screens an orthographic projection view to identify manipulable parts coarsely.
We then annotate and calibrate each part's joint type, valid motion range, and manipulation semantics, enabling agentic task generation (Sec.~\ref{subsec:task_generation}) to select semantically consistent actions.
DINO-X~\citep{ren2024dino} then localizes each part in the 2D view, and ray tracing recovers its corresponding surface point coordinates in the object frame.
From these predictions, an anchor action frame is extracted through iterative erosion and 2D-to-3D back-projection.
Based on the joint type, rule-based sampling around the anchor produces a set of feasible action frames that introduce interaction diversity for motion generation.
Finally, the VLM identifies part materials to support fine-grained, part-wise and joint-wise visual and physical randomization (Sec.~\ref{subsec:task_domain_randomization}).
The resulting part and joint annotations provide structured action targets for downstream manipulation tasks.
See Fig.~\ref{app:subfig:arti_anno} for examples of the resulting annotations.

\textbf{Motion generation.}
We generate trajectories based on the annotated action frames and robot states.
In collision-free space, we use cuRobo~\citep{sundaralingam2023curobo} for GPU-accelerated batch trajectory optimization and collision-aware motion planning in joint space, and we perturb the resulting trajectories to introduce motion variations.
For in-contact interactions, motion generation is subject to tighter constraints, as the end-effector must maintain contact while following the articulation constraints throughout the interaction.
To address this, we leverage simulation privileged information, including the exact joint types, positions, and valid motion ranges at runtime, to formulate trajectory optimization with these quantities as explicit constraints (see App.~\ref{app:subsec:constraint_solver} for details).
The resulting trajectories conform to the underlying articulation constraints, improving interaction stability in simulation.
Since variations in action frames may introduce unstable or infeasible motions, we leverage the high parallelism of \ourframework to efficiently filter out low-quality trajectories.
Together, these designs support diverse articulated interactions while maintaining stable and high-quality motions.

\textbf{Manipulation skills.}
To enable efficient agentic task planning, we group joint types and manipulation semantics into 23 manipulation skills.
These skills abstract low-level motion primitives into reusable, semantically specified actions that the agent can select and compose when generating tasks.
Each skill is tailored with specific constraint types and parameterizations according to its typical motion pattern and grasping mode, enabling better adaptation across objects and task configurations.

The resulting skill set covers translation-based interactions (e.g., pushing, pulling, sliding, and pressing), rotation-based interactions (e.g., opening a door, turning, flipping, and screw/knob actuation),
and pick-and-place behaviors (e.g., picking, placing, and aligning).
Each skill supports decomposition of a manipulation process into sub-processes whose motions are generated through sampling-based planning or constraint-guided trajectory generation, depending on the interaction structure and task requirements.
These sub-processes are parameterized so that the same skill can be adapted across different object annotations and task requirements.
The skills serve as reusable building blocks for composing complex manipulation behaviors.
To improve robustness and diversity, each skill further supports randomized perturbations of control points, introducing trajectory-level variations while preserving its semantic intent.
During agentic planning for articulated-object manipulation, skills are selected based on the manipulation semantics of the target parts, enabling fine-grained manipulation behaviors and more effective motion generation.

\textbf{Environment parallelism.}
To enable scalable demonstration collection, we leverage Isaac Lab's parallel simulation and implement an asynchronous plan-and-execute framework with tensor-based action assembly.
To handle variable trajectory lengths induced by randomization and planning initialization, we decouple motion generation from action assembly, enabling independent trajectory construction across different task stages. 
This design enables massively parallel environment execution and maximizes runtime utilization.

\subsection{Task Domain Randomization}
\label{subsec:task_domain_randomization}

\begin{figure*}[t]
    \centering

    \begin{subfigure}[b]{\textwidth}
        \centering
        \includegraphics[width=0.235\linewidth]{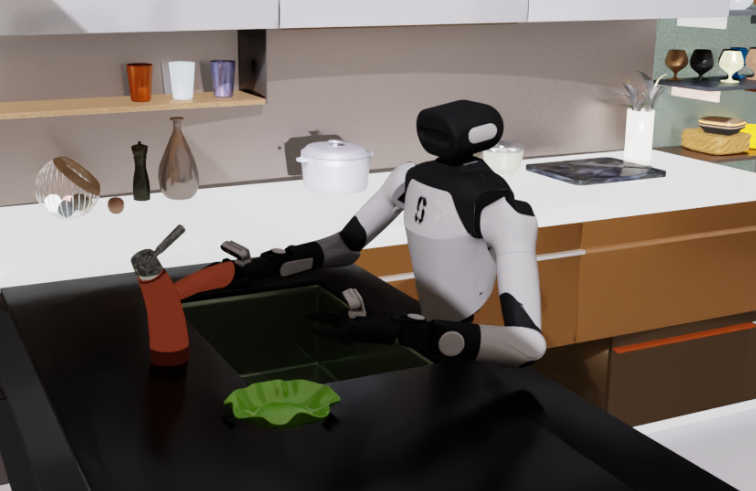}
        \hfill
        \includegraphics[width=0.235\linewidth]{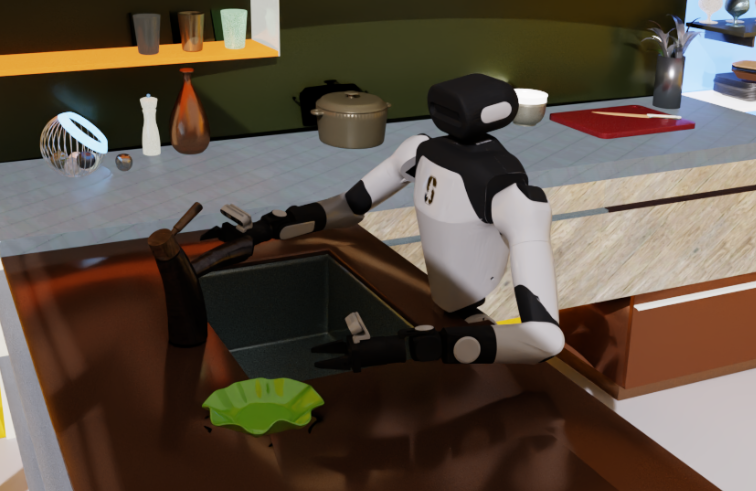}
        \hfill
        \includegraphics[width=0.235\linewidth]{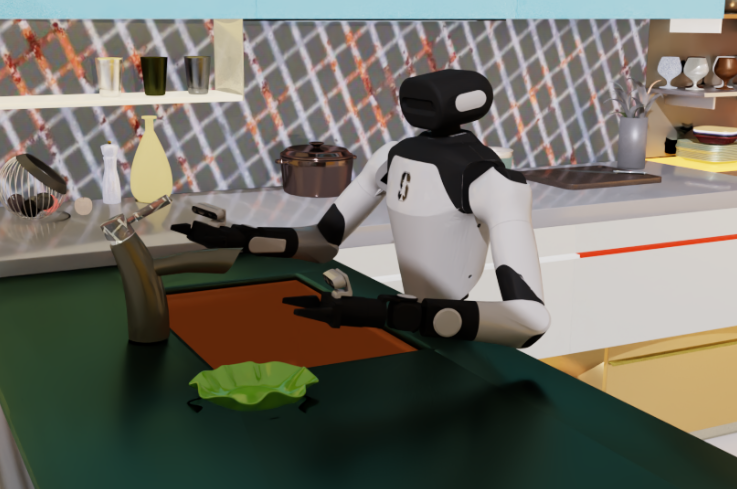}
        \hfill
        \includegraphics[width=0.235\linewidth]{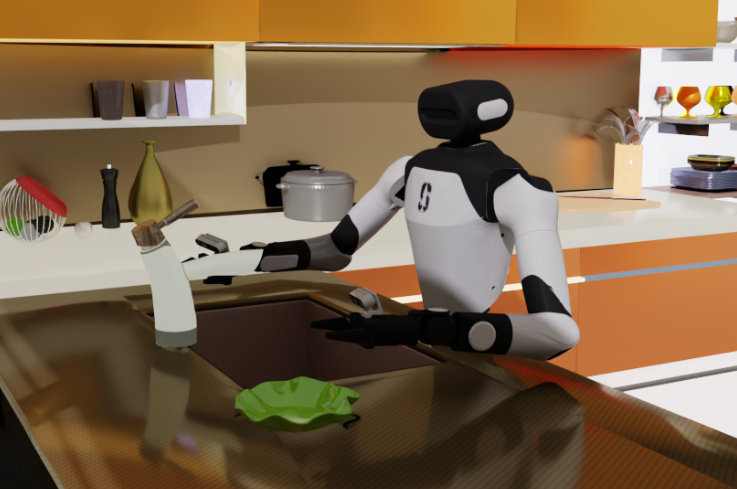}
        \caption{Visualization of visual material randomization.}
        \label{fig:vis_rand}
    \end{subfigure}

    \medskip

    \begin{subfigure}[b]{\textwidth}
        \centering
        \includegraphics[width=0.235\linewidth]{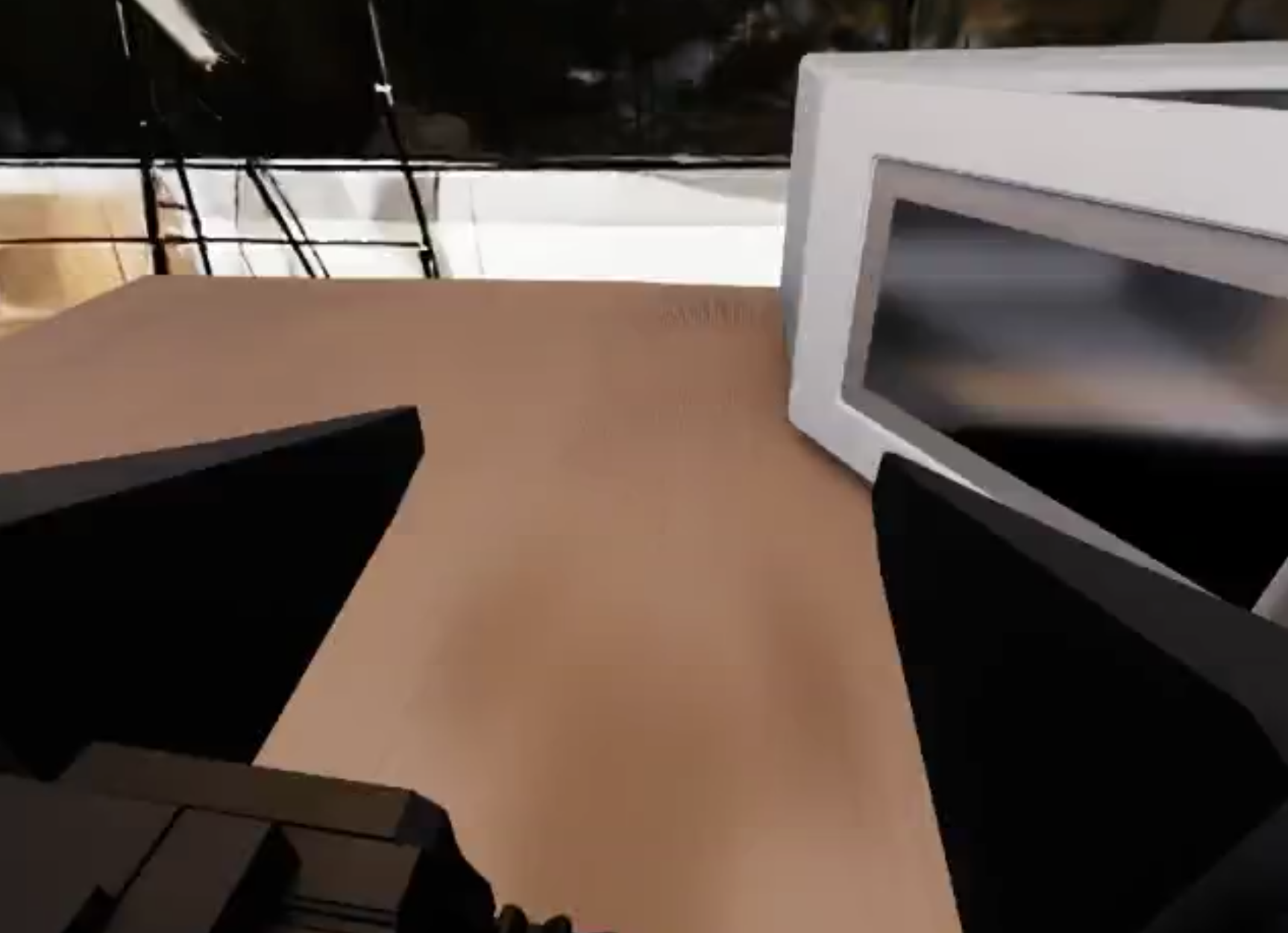}
        \hfill
        \includegraphics[width=0.235\linewidth]{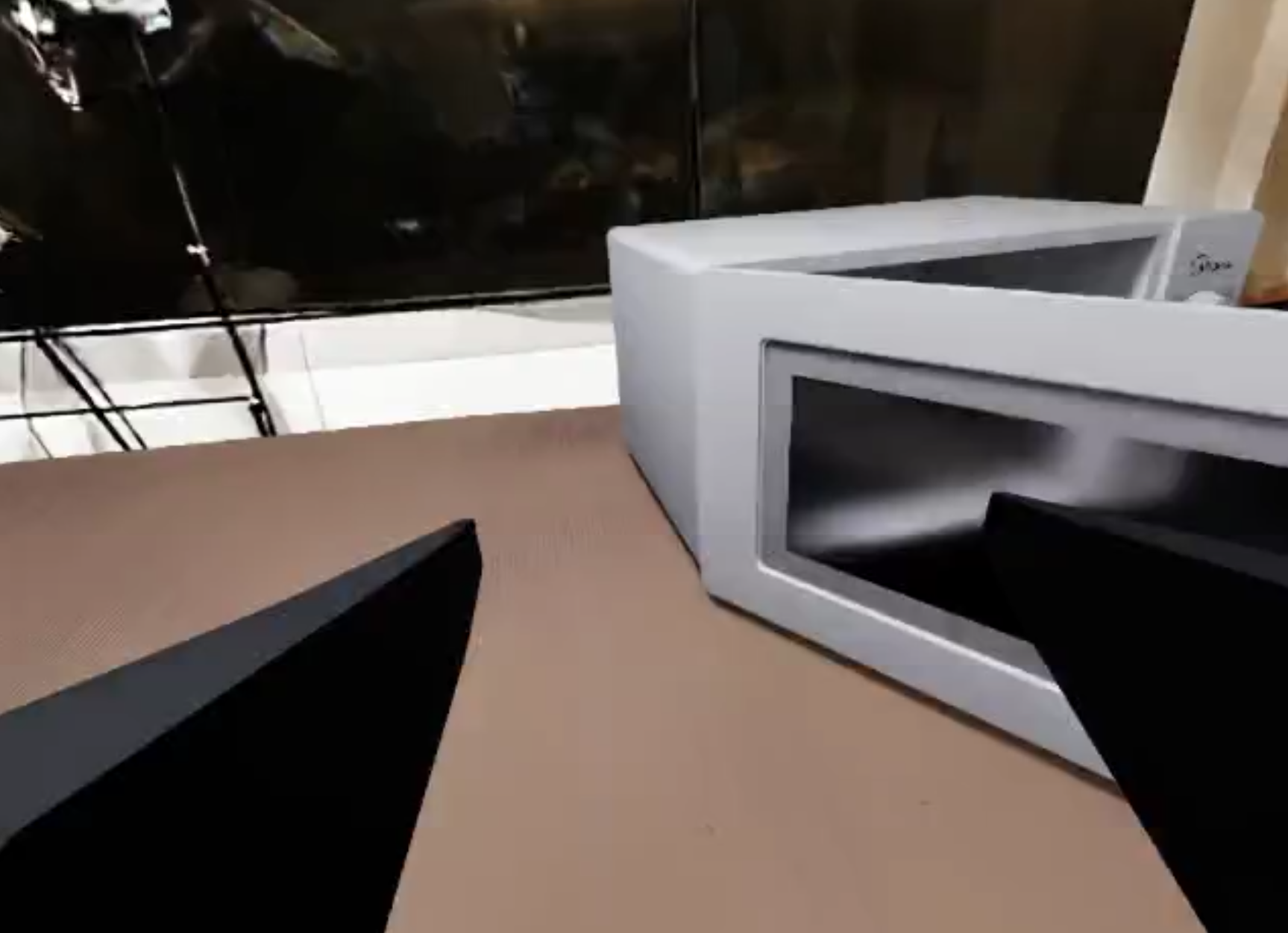}
        \hfill
        \includegraphics[width=0.235\linewidth]{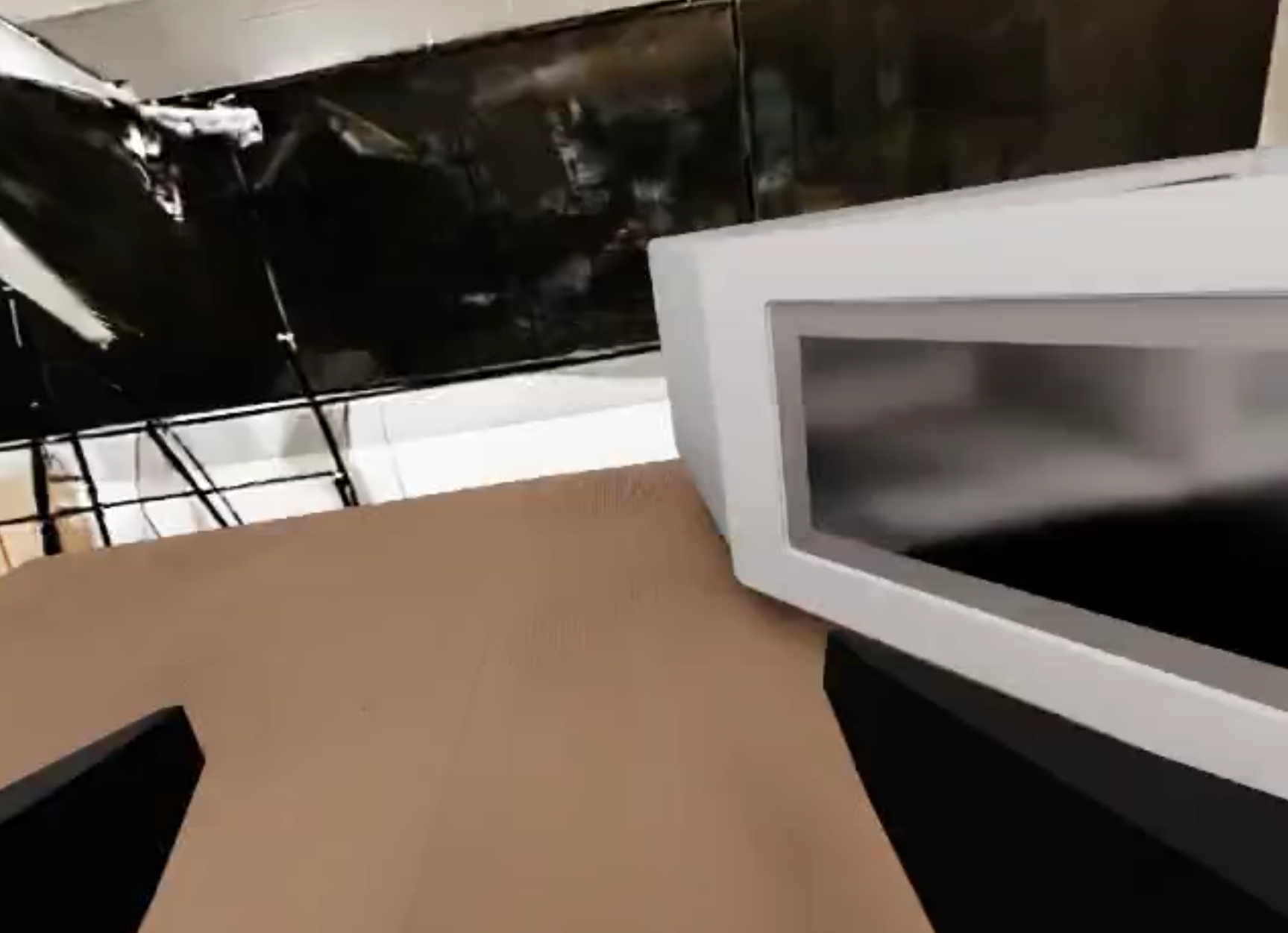}
        \hfill
        \includegraphics[width=0.235\linewidth]{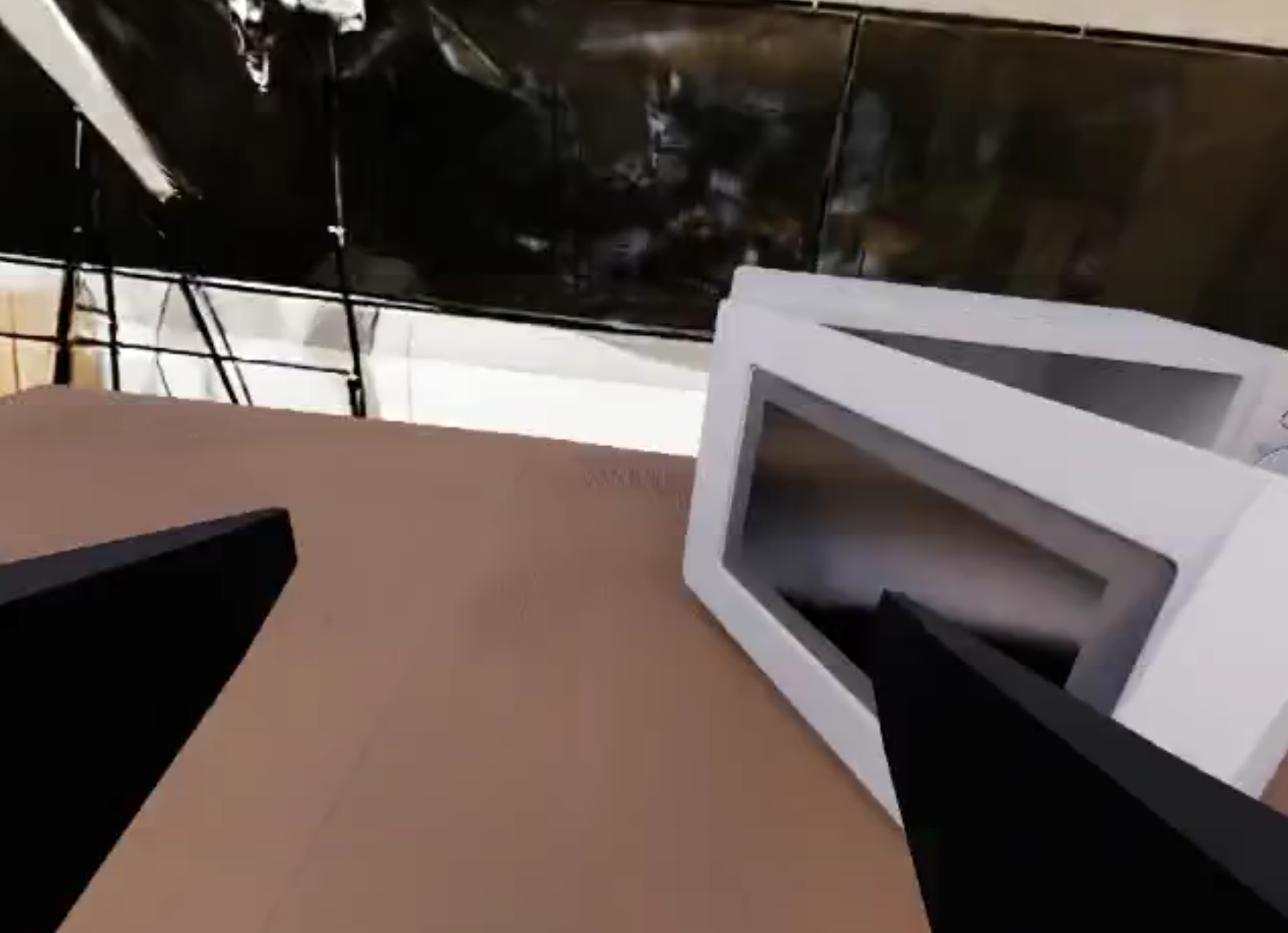}
        \caption{Visualization of camera randomization.}
        \label{fig:cam_rand}
    \end{subfigure}

    \medskip

    \begin{subfigure}[b]{\textwidth}
        \centering
        \includegraphics[width=0.235\linewidth]{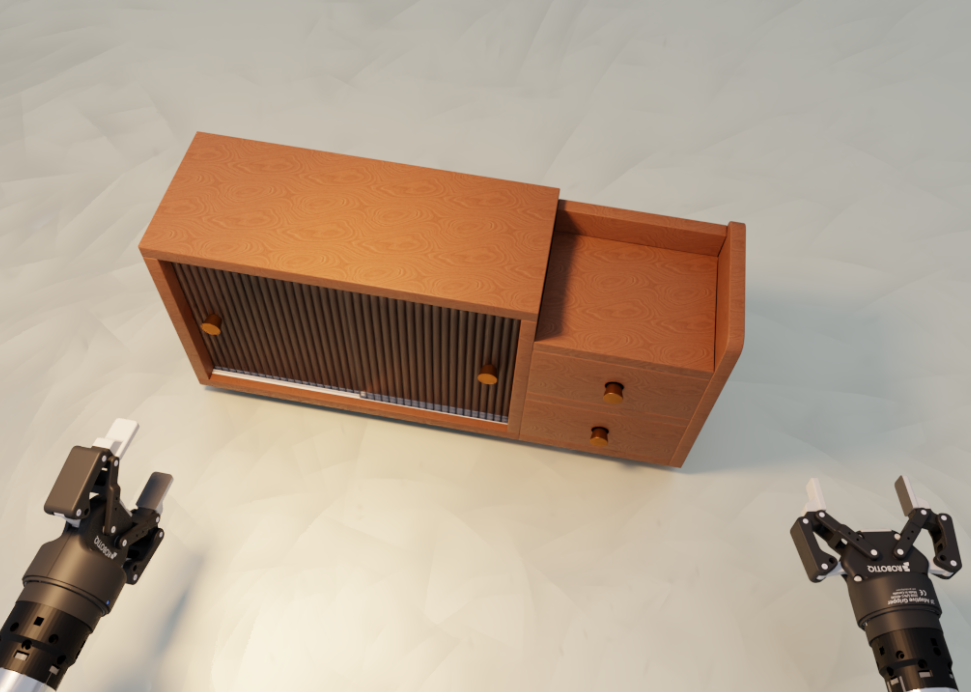}
        \hfill
        \includegraphics[width=0.235\linewidth]{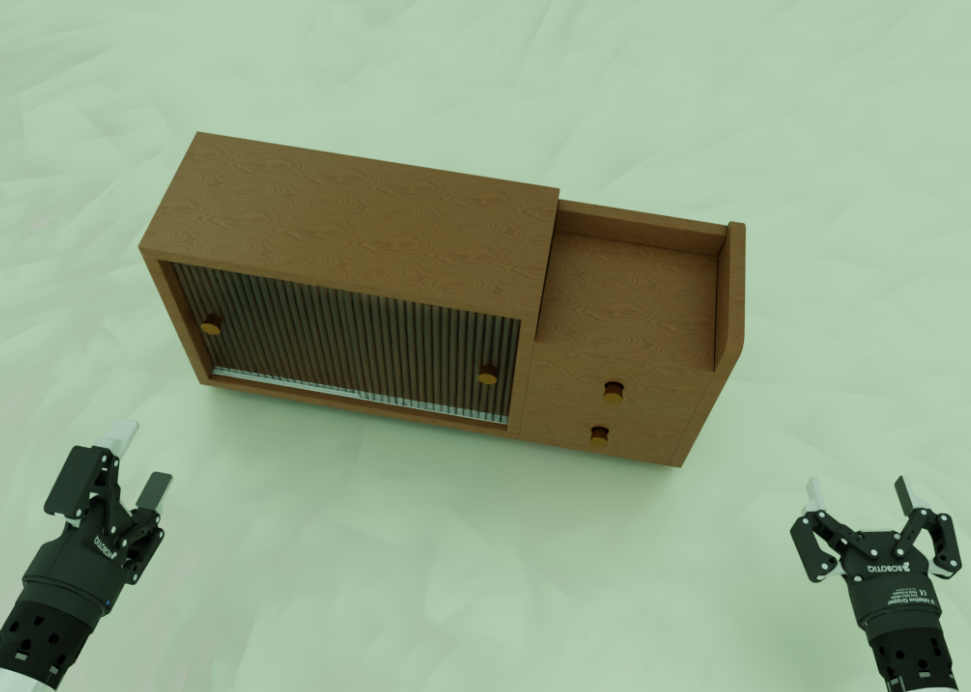}
        \hfill
        \includegraphics[width=0.235\linewidth]{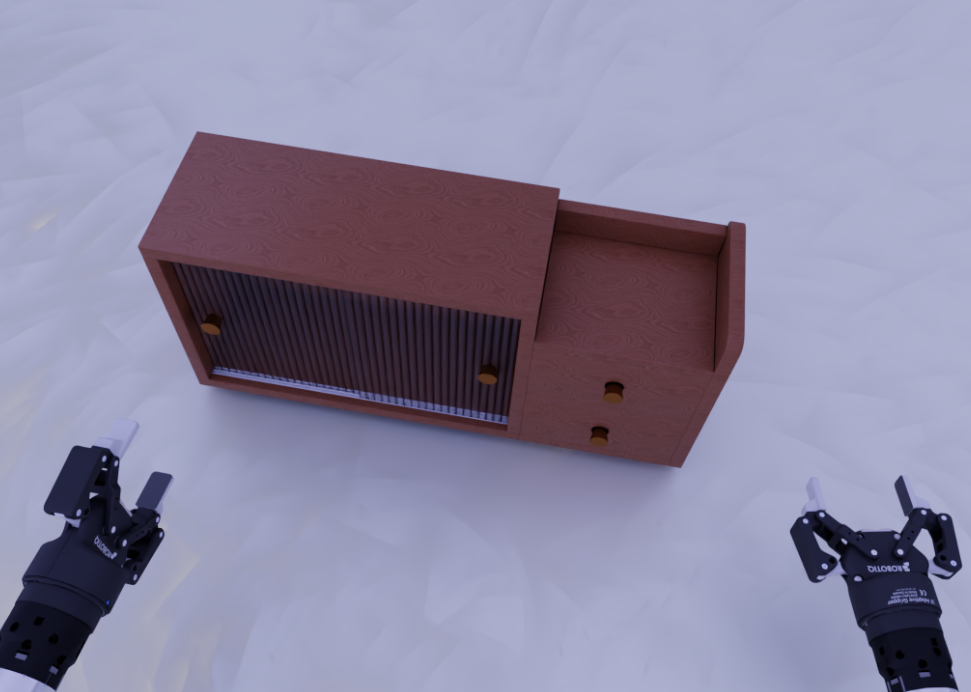}
        \hfill
        \includegraphics[width=0.235\linewidth]{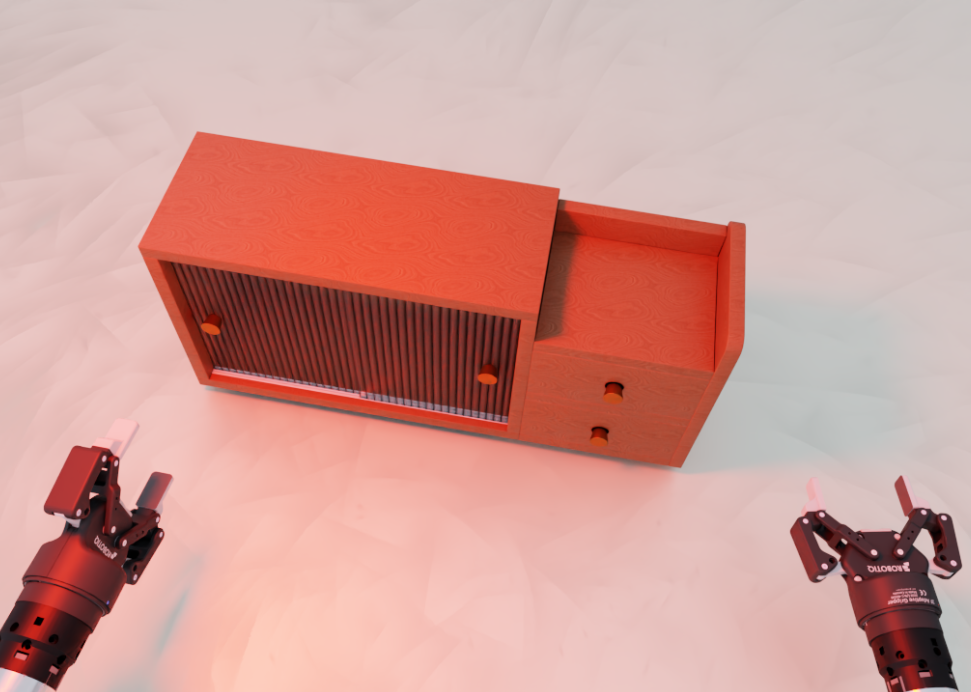}
        \caption{Visualization of illumination randomization.}
        \label{fig:illum_rand}
    \end{subfigure}
    
    \medskip
    \begin{subfigure}[b]{\textwidth}
        \centering
        \includegraphics[width=0.235\linewidth]{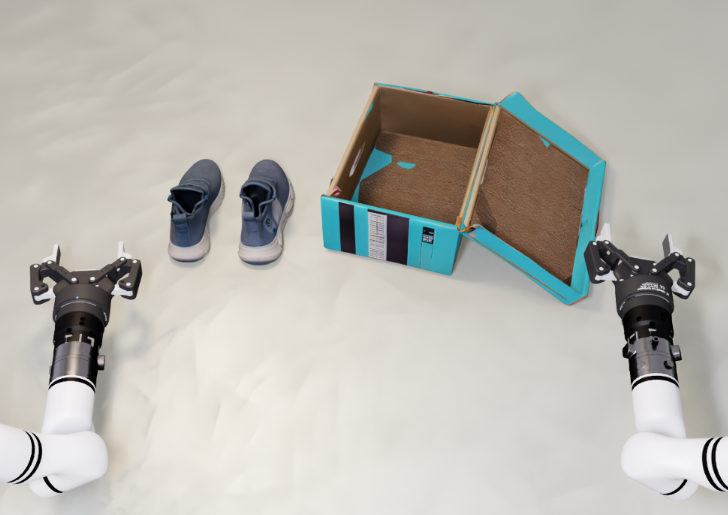}
        \hfill
        \includegraphics[width=0.235\linewidth]{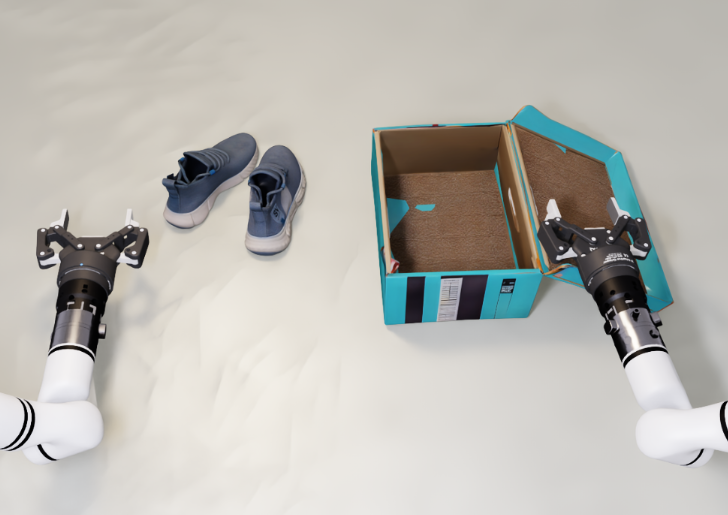}
        \hfill
        \includegraphics[width=0.235\linewidth]{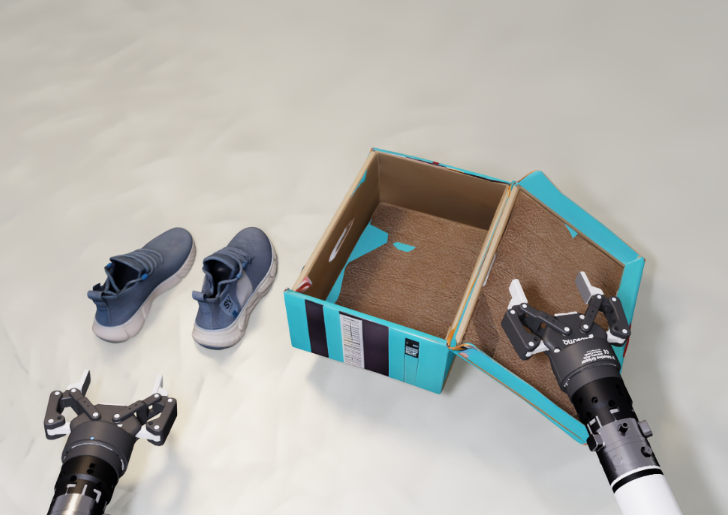}
        \hfill
        \includegraphics[width=0.235\linewidth]{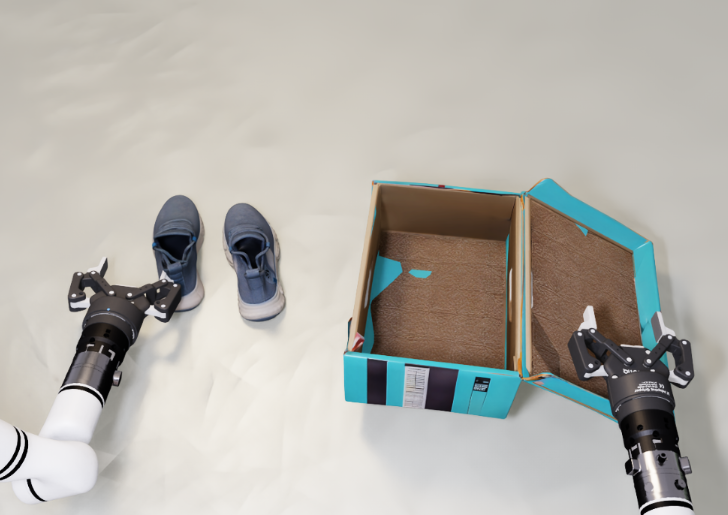}
        \caption{Visualization of spatial and robot configuration randomization.}
        \label{fig:spatial_rand}
    \end{subfigure}

    \caption{The articulation-aware domain randomization effects supported by \ourframework.}
    \label{fig:domain_randomization}
\end{figure*}

\ourframework introduces articulation-aware designs for domain randomization to increase demonstration diversity and vary the conditions represented in the synthesized data.
These designs follow the multi-part structure, functional semantics, and physical constraints of articulated objects, moving beyond coarse object-level randomization.
The resulting randomization covers spatial, physical, and visual aspects, as illustrated in Fig.~\ref{fig:domain_randomization}.

\textbf{1) Spatial randomization.}
In each episode, we randomize object poses, robot initial configurations, and camera extrinsics.
For articulated objects that contain other interactable objects, we introduce a group randomization mechanism that first transforms the container and contained objects as a group and then independently randomizes each contained object, while perturbing their relative poses within predefined ranges (Fig.~\ref{fig:spatial_rand}).
For robot configurations, the initial joint state is sampled around well-tuned nominal configurations to vary feasible initial states across episodes.
For camera configurations, intrinsic parameters are aligned with the target camera, while extrinsics are sampled within ranges that represent potential mounting errors, calibration noise, and slight pose deviations in real-world setups (Fig.~\ref{fig:cam_rand}).
These procedures diversify object placements, inter-object geometry, robot states, and camera viewpoints across synthesized episodes.

\textbf{2) Physical randomization.}
Articulated objects consist of multiple movable parts with distinct physical properties, making part-wise randomization necessary for representing dynamics variations.
We therefore independently randomize the friction and density of each movable part, while perturbing joint stiffness and damping within small ranges.
Introducing part-level and joint-level randomization produces richer variation in contact and articulation dynamics, exposing the policy to diverse physical interactions across movable parts and their joints.

\textbf{3) Visual randomization.}
Articulated objects consist of multiple parts with distinct functional semantics, and coarse object-level randomization may obscure these part-level visual distinctions.
To diversify visual appearances while preserving the semantic structure of each object in visual observations, we adopt fine-grained, part-level randomization by independently sampling materials and textures for different components from pre-designed libraries (Fig.~\ref{fig:vis_rand}).
We further randomize local and global illumination by perturbing light placement, intensity, color, and color temperature, and by sampling environment maps from an HDR environment library (Fig.~\ref{fig:illum_rand}).
These settings create variation in direct, ambient, and reflected illumination across synthesized episodes.

The sampling ranges for all randomization aspects are provided in App.~\ref{app:subsec:randomization_details}.

%% file: tables/framework_comparison.tex
\begin{table}[H]
\centering
\scriptsize
\setlength{\tabcolsep}{3.2pt}
\renewcommand{\arraystretch}{0.95}
\caption{
Comparison of common robot simulation frameworks. 
Some frameworks only list the total number of objects and do not distinguish between rigid and articulated objects.
}
\resizebox{\textwidth}{!}{
\begin{tabular}{
l
>{\columncolor{ourblue}}c
cccccccccc
}
\toprule
\textbf{Feature}
& \makecell{\textbf{\ourframework}\\(Ours)}
& \makecell{\textbf{Robo-}\\\textbf{Casa365}}
& \makecell{\textbf{Molmo-}\\\textbf{Space}}
& \makecell{\textbf{Arti-}\\\textbf{Bench}}
& \makecell{\textbf{RoboTwin}\\\textbf{2.0}}
& \makecell{\textbf{RL-}\\\textbf{Bench}}
& \makecell{\textbf{Behavior-}\\\textbf{1K}}
& \makecell{\textbf{Humanoid-}\\\textbf{Gen}}
& \makecell{\textbf{Mani-}\\\textbf{Skill 2}}
& \textbf{LIBERO}
& \makecell{\textbf{InternData-A1}} \\
\midrule
Scenes            & 1{,}452 & 120  & 230{,}000+  & 1 & 1 & 1   & 50   & 20 & --   & 20  & 227 \\
Embodiments       & 7 & 1 & 2  & 1  & 5  & 1  & 12 & 1  & 1 & 1  & 4 \\
(Rigid) Objects           & 9{,}116 & 2{,}509 & 130{,}000+  & --  & 687 & 28  & 9318 & --  & 2144 & --  & 3185 \\
Articulated Objects & 2{,}507 & 20 & -- & -- & 44 & -- & -- & 4 & -- & -- & 321 \\

\midrule
Realistic Physics       & \cmark & \cmark & \cmark & \cmark & \cmark & \cmark & \cmark & \cmark & \cmark & \cmark & \cmark \\
Realistic Rendering     & \cmark & \xmark & \xmark & \xmark & \cmark & \xmark & \cmark & \cmark & \cmark & \xmark & \cmark \\
Object Annotation       & \cmark & \xmark & \cmark & \cmark & \cmark & \xmark & \xmark & \cmark & \xmark & \xmark & \cmark \\
Scripted Datagen        & \cmark & \cmark & \cmark & \cmark & \cmark & \cmark & \xmark & \cmark & \cmark & \xmark & \cmark \\
AI-generated Tasks      & \cmark & \xmark & \cmark & \xmark & \xmark & \xmark & \xmark & \cmark & \xmark & \xmark & \cmark \\
Multi-environment Parallelism & \cmark & \xmark & \xmark & \xmark & \xmark & \xmark & \xmark & \xmark & \cmark & \xmark & \xmark \\
Distributed Data Synthesis & \cmark & \xmark & \xmark & \xmark & \xmark & \xmark & \xmark & \xmark & \xmark & \xmark & \cmark \\
\bottomrule
\end{tabular}
}
\label{tab:sim-framework-comparison}
\end{table}

%% file: sections/04_data_synthesis.tex
\section{Data Synthesis}
\label{sec:data_synthesis}

\subsection{Agentic Task Generation}
\label{subsec:task_generation}

We implement a task generation agent to convert natural-language task descriptions into executable simulation configurations (Fig.~\ref{fig:data_syn}), with the following components:

\begin{figure}[h]
    \centering
    \includegraphics[width=1\linewidth]{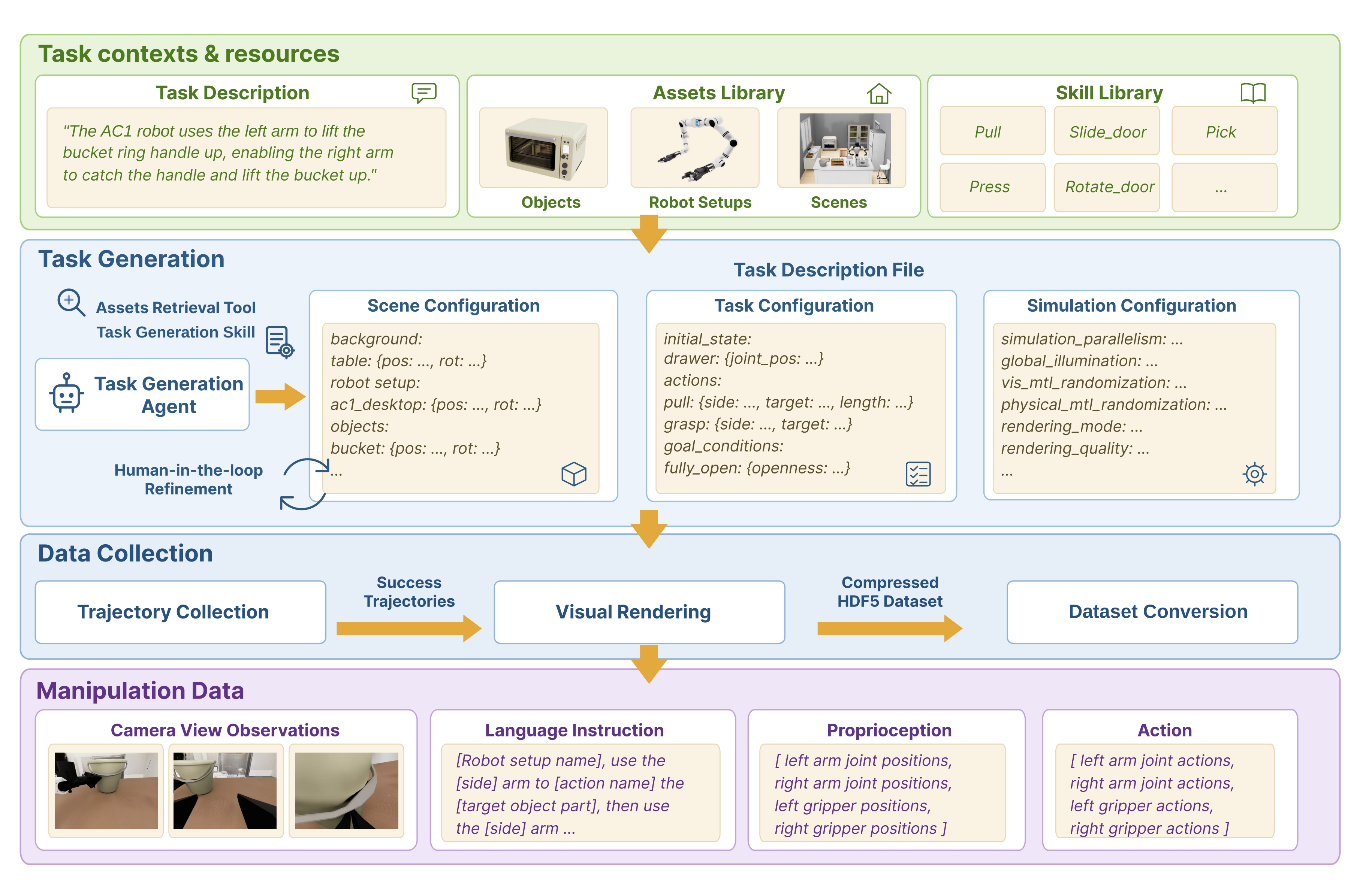}
    \caption{Illustration of our agentic task generation and distributed data synthesis.}
    \label{fig:data_syn}
\end{figure}

\textbf{1) Unified task specification.}
Inspired by task specification languages such as BDDL~\citep{li2023behavior1k}, we introduce a unified task format covering environment configuration (scenes, objects, and robot embodiments), task definition (skill sequences, initial states, and success conditions), and simulation settings (rendering, physics, randomization, and parallelization), bridging high-level task intent and low-level task execution. This structured representation serves as explicit context for the LLM, enabling it to translate abstract instructions into grounded, executable simulator configurations.
Environment configuration describes the scene layout, background objects, foreground workspace, task-relevant objects, and robot embodiment.
Task definition specifies the skill sequence, logical initial conditions, manipulation procedure, target goal, and success criterion.
Simulation settings specify execution parameters, including rendering options, physical properties, domain randomization, and parallel execution settings.
See App.~\ref{app:subsec:task_file} for an example.

\textbf{2) Scene grounding and skill program generation.}
Given a task description, the agent first retrieves the required robot, objects, and scene from the resource library using an asset retrieval tool. 
At initialization, the platform provides a task construction library containing available robot setups, rigid and articulated object assets, and background scenes.
Using the asset catalog and in-context examples of valid configurations, the agent parses the task description into the required robot setup, target manipulation objects, scene type, and auxiliary objects, and grounds these elements to simulator assets.
The agent then determines the robot configuration, object identities, initial object states, and object poses, producing a physically plausible and task-relevant scene configuration that conforms to simulator constraints.
It then leverages its predefined task generation skill, querying the manipulation annotations of the target object part (Sec.~\ref{subsec:manipulation_annotation}) and translating task semantics into a parameterized program of skills introduced in Sec.~\ref{subsec:motion_gen}. 
The manipulation skills used in this program are the 23 reusable skills already defined in Sec.~\ref{subsec:motion_gen}, where \ourframework groups joint types and manipulation semantics and tailors each skill with specific constraint types and parameterizations according to its motion pattern and grasping mode.
Guided by the task semantics and part annotations, the agent selects and composes these existing skills into a parameterized high-level skill program.

\textbf{3) Simulation configuration setting.}
The agent then configures the simulation settings according to task contents and expected simulation overhead, including domain randomization described in Sec.~\ref{subsec:task_domain_randomization}, rendering mode, parallelism, etc. 
Lighting presets are selected to vary illumination conditions, while rendering settings are chosen according to object material properties.
For example, tasks involving transparent or reflective objects can be assigned higher-quality rendering modes such as path tracing~\citep{ouyang2021restir}.
The number of parallel environments is adjusted according to the expected task horizon estimated from the generated skill sequence, balancing generation throughput against memory consumption.
Human-in-the-loop refinement can optionally be performed through the simulator UI.


\subsection{Data Collection} 
\label{subsec:data_collection}

Efficient large-scale demonstration generation is bottlenecked by rendering computation wasted on trajectories that ultimately fail and by the mismatch between CPU-bound planning and GPU-intensive rendering.
Inspired by Nimbus~\citep{he2026nimbus}, we address this with a distributed system consisting of a scheduler and three workload types:

\textbf{1) Trajectory collection.}
Workers run environments in parallel, retain only successful trajectories, and apply quality-control conditions to trigger early resets.
During trajectory collection, task-level parallelism can be set up with up to 500 environments, and all environments share one motion planner to reduce GPU memory consumption.
Jerk and configuration-deviation checks trigger early resets for unstable executions, so only successful trajectories enter rendering.
This stage produces approximately 1.25 trajectories per second.

\textbf{2) Visual rendering.}
Workers replay successful trajectories using their stored seeds to ensure deterministic motion replay.
Visual rendering uses tiled cameras in Isaac Lab for parallel multi-view rendering and incrementally encodes HDF5 data during simulation to avoid end-of-episode CPU spikes.

\textbf{3) Dataset conversion.}
Workers convert the rendered HDF5 files into the LeRobot~\citep{cadene2026lerobot} format with redundant fields filtered and streaming video encoding, running conversion alongside collection and rendering.
Dataset conversion applies key mapping across robot embodiments and uses per-frame stream encoding for multi-camera video.

Trajectory collection and visual rendering share one GPU pool, while dataset conversion runs on CPUs and can continue alongside the GPU-bound stages.
All stages communicate through a shared storage pool, so intermediate HDF5 products can be consumed immediately by the next stage without explicit data movement.
The scheduler uses first-in-first-out dispatch by default and supports priority scheduling for urgent tasks.
It can split a task with many episodes into independent subtasks to limit peak storage use, and the command-line interface reports task status and resets failed tasks.
The released \ourdataset was generated with 96 workers in total.
With eight RTX 4090 GPUs, one server generates approximately 144 hours of data per day.
By decoupling these CPU- and GPU-intensive stages, the system is highly scalable with more workers.

\subsection{Dataset Statistics}
\label{subsec:dataset_statistics}

\begin{figure}[t]
    \centering
    \begin{subfigure}[t]{0.26\linewidth}
        \centering
        \includegraphics[width=\linewidth, valign=t]{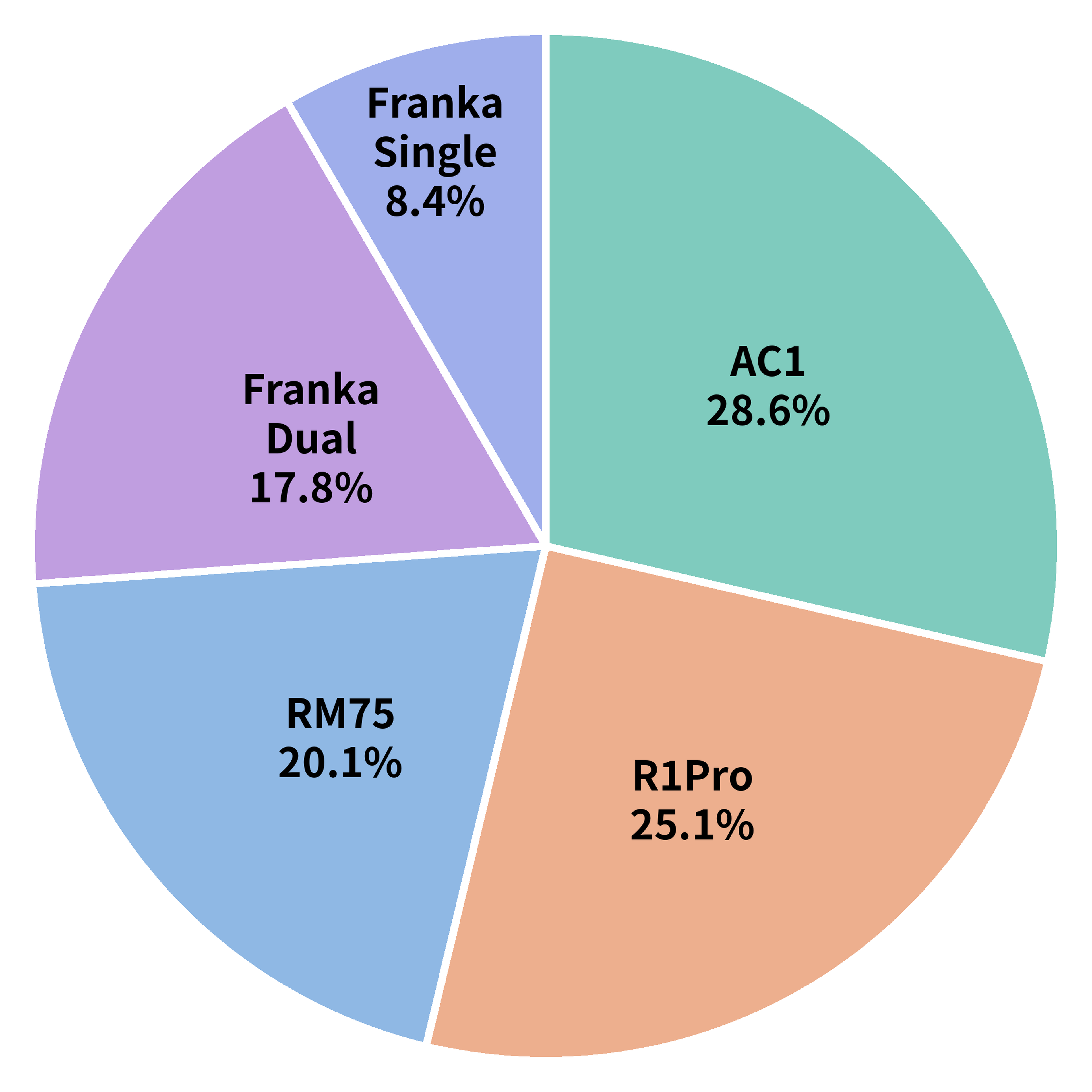}
        \caption{Distribution of trajectories across five robot setups.}
        \label{subfig:dist_robot_setup}
    \end{subfigure}
    \hfill
    \begin{subfigure}[t]{0.26\linewidth}
        \centering
        \includegraphics[width=\linewidth, valign=t]{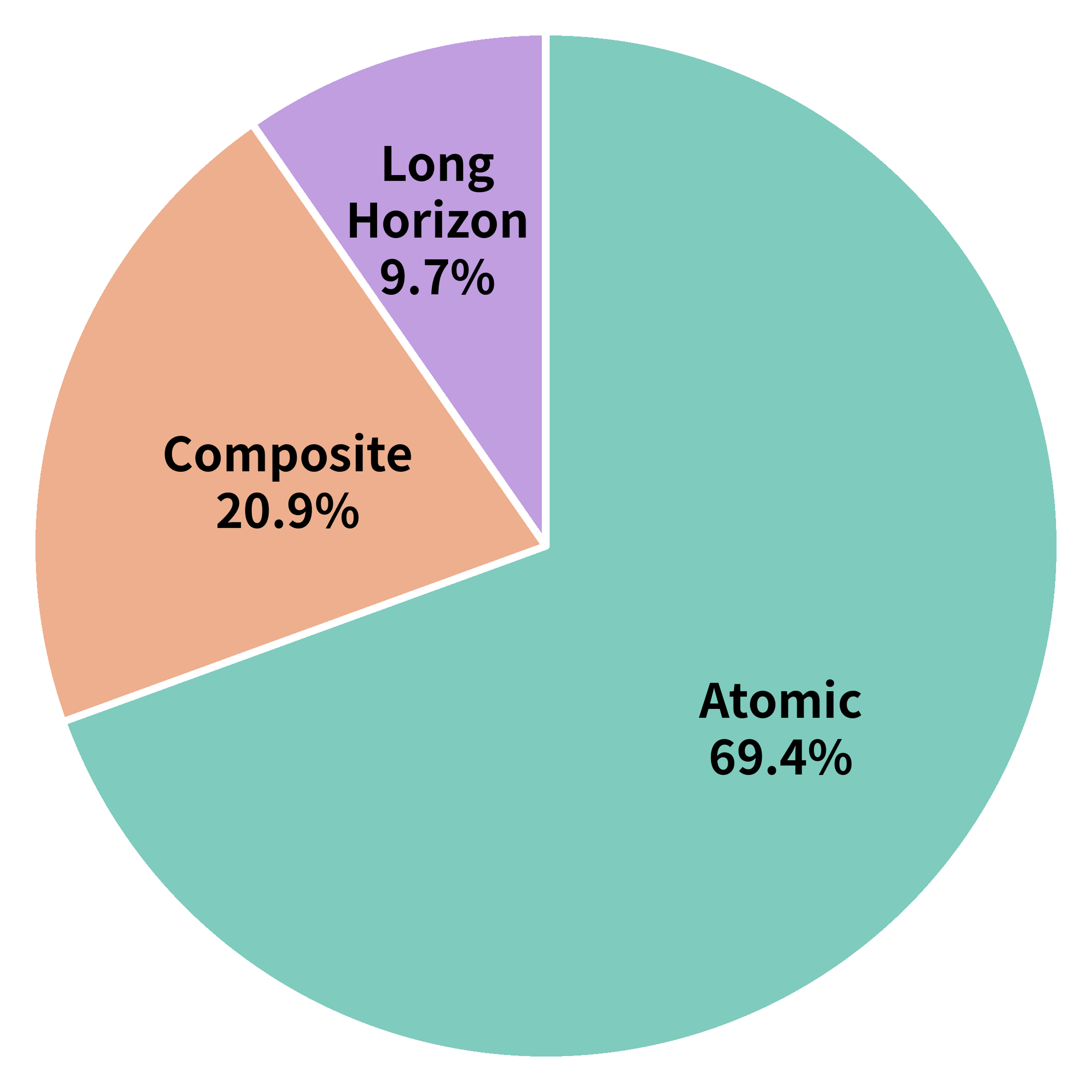}
        \caption{Distribution of trajectories over three task complexity levels.}
        \label{subfig:dist_task_complexity}
    \end{subfigure}
    \hfill
    \begin{subfigure}[t]{0.45\linewidth}
        \centering
        \includegraphics[width=\linewidth, valign=t]{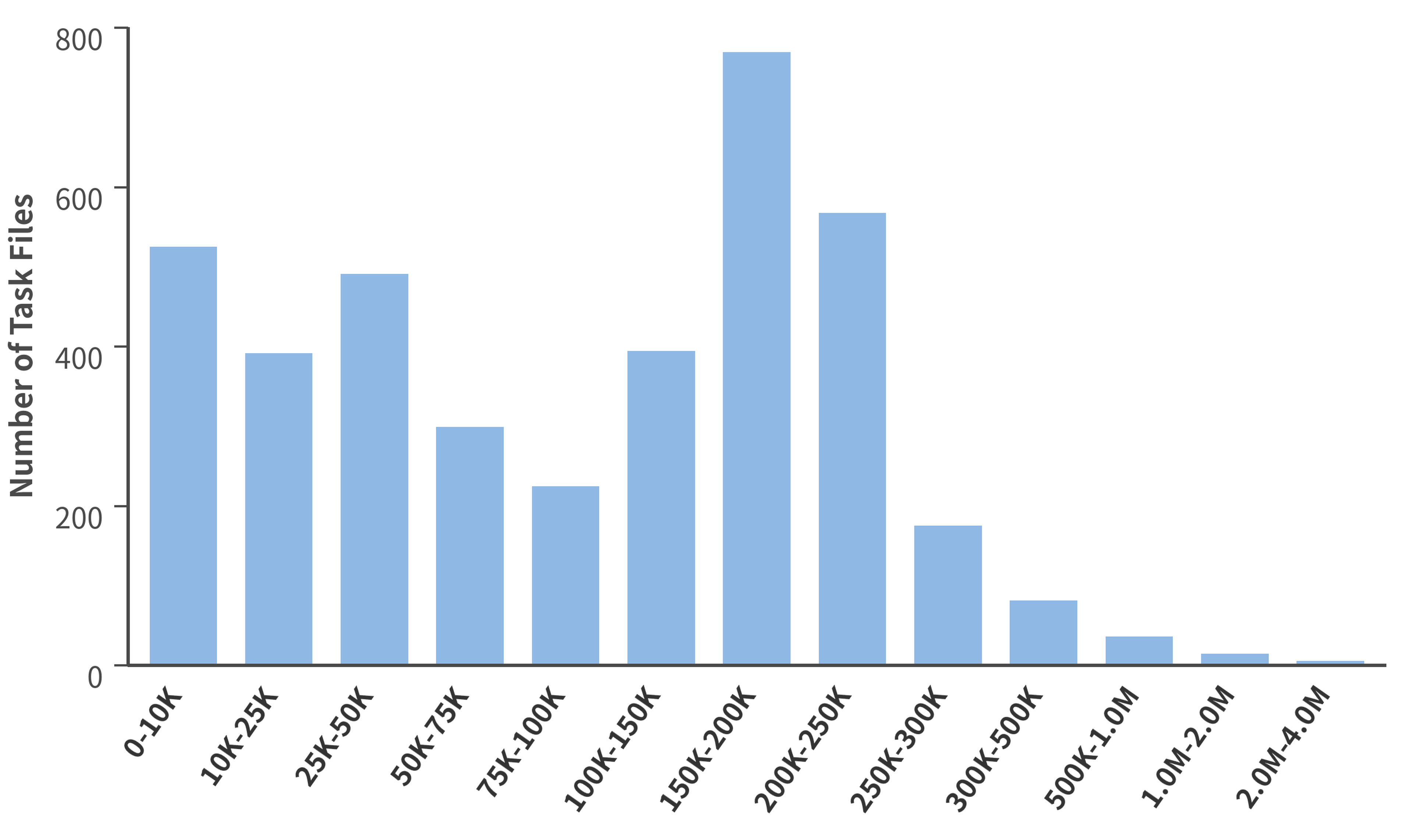}
        \caption{Distribution of task files over task durations.}
        \label{subfig:dist_task_duration}
    \end{subfigure}

    \begin{subfigure}[t]{0.49\linewidth}
        \centering
        \includegraphics[width=\linewidth, valign=t]{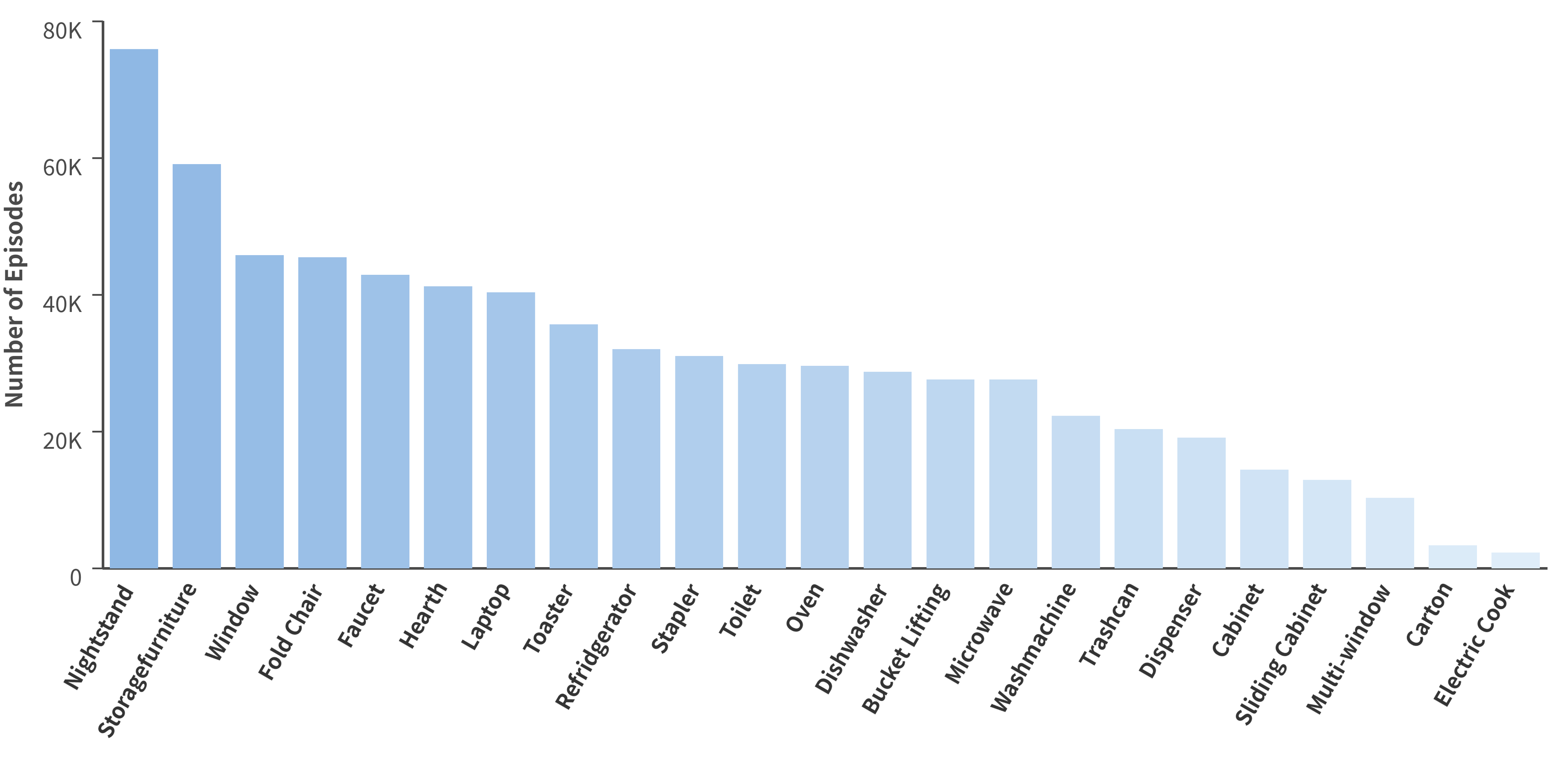}
        \caption{Distribution of trajectories over articulated object categories.}
        \label{subfig:dist_articulation_category}
    \end{subfigure}
    \hfill  
    \begin{subfigure}[t]{0.5\linewidth}
        \centering
        \includegraphics[width=\linewidth, valign=t]{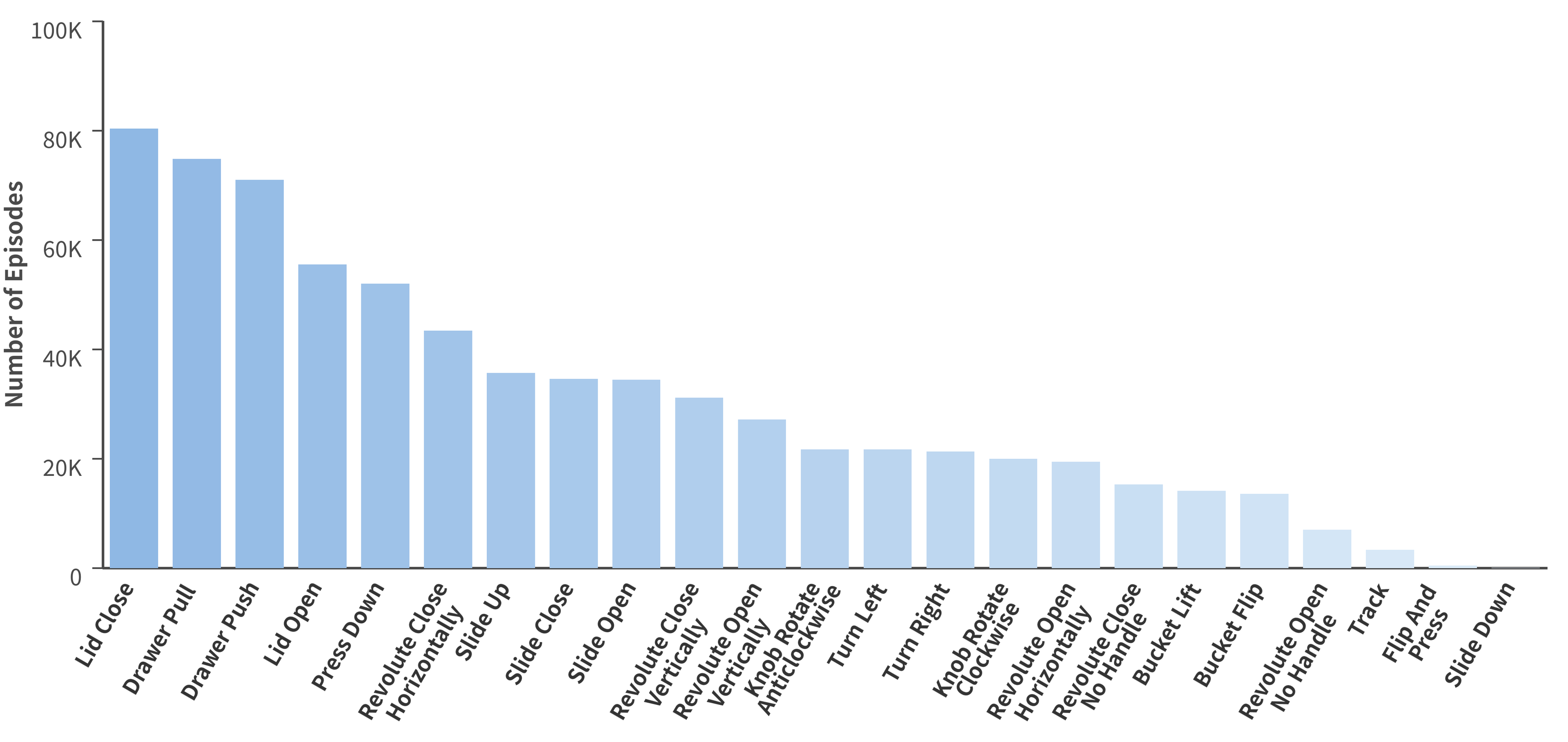}
        \caption{Distribution of trajectories over manipulation skills.}
        \label{subfig:dist_skill}
    \end{subfigure}

    \caption{Statistics of \ourdataset over robot setups, task complexity levels, skills, articulated object categories, and task durations.}
    \label{fig:dataset_stats}
\end{figure}

Using the data production approach above, we construct \textbf{SMART-Data}, a large-scale, high-quality dataset for articulated-object manipulation.
It comprises over 1M episodes collected across five robot setups, totaling approximately 500M frames and over 4,600 hours of manipulation data, with 23 manipulation skills, 44 atomic task types, and 1{,}122 distinct scenes.
Fig.~\ref{fig:dataset_stats} summarizes its main statistics across robot embodiments, task categories, skills, task durations, and articulated object categories.
As shown in Tab.~\ref{tab:dataset_comparison}, \ourdataset provides the largest articulated-object trajectory set as well as broad skill, scene, and embodiment coverage among the compared datasets.
This scale and diversity provide sufficient training data for robot learning models and support comprehensive evaluation across diverse scenarios.

\input{tables/dataset_comparison}

\textbf{Tasks.}
Our task design implements a hierarchical taxonomy spanning multiple complexity levels, and the released trajectories comprise 69.4\% atomic tasks, 20.9\% composite tasks, and 9.7\% long-horizon tasks (Fig.~\ref{subfig:dist_task_complexity}).
The dataset encompasses 23 distinct manipulation skills, 44 atomic task types, and standardized definitions across 10k task YAML configurations, and the skill distribution is shown in Fig.~\ref{subfig:dist_skill}.
The durations of task files span a wide range, from a few hundred to several million frames, as shown in Fig.~\ref{subfig:dist_task_duration}.
The task-file duration distribution indicates substantial long-horizon coverage, while the skill distribution shows that the synthesized trajectories are broadly distributed across manipulation skills rather than concentrated on a small subset.

\textbf{Robots.}
The dataset spans five robot setups constructed from the embodiments supported by \ourframework: AC1, R1Pro, Dual RM75, Dual Franka, and Single Franka, comprising 1,004,584 trajectories whose distribution is shown in Fig.~\ref{subfig:dist_robot_setup}.
Detailed robot platform specifications are provided in App.~\ref{app:subsec:robot_platform_details}.

\textbf{Object diversity.}
\ourdataset incorporates the 23 functional categories that appear in the collected trajectories out of the 88 categories available in \ourframework's full asset library (Sec.~\ref{subsec:simulation_assets}), including cabinets, drawers, windows, refrigerators, ovens, dishwashers, and dispensers with diverse joint mechanisms, as shown in Fig.~\ref{subfig:dist_articulation_category}.
These articulated objects exhibit varied kinematic structures such as revolute joints, prismatic joints, and compound mechanisms, requiring different manipulation strategies. 
Object diversity extends beyond geometry to include rich texture variation, with procedurally generated materials and realistic surface properties applied to object surfaces.

%% file: tables/dataset_comparison.tex
\begin{table}[H]
\centering
\caption{Comparison of robotic simulation datasets with substantial articulated-object manipulation data. 
For \ourdataset, we only report the number of atomic tasks, as the high diversity of composite tasks makes a consistent quantitative comparison challenging.
}
\label{tab:dataset_comparison}
\vspace{-2mm}

\large
\resizebox{\textwidth}{!}{
\begin{tabular}{lccccccc}
\toprule
Dataset 
& Traj. (All)
& Traj. (Arti.)
& Skill 
& Task 
& Scene 
& Embodiment 
& Collection Method\\
\midrule

RoboCasa365
& 655k 
& 415k 
& 8 
& 100 
& 120 
& 1 
& Teleoperation \& Augmentation \\

RoboTwin 2.0
& 100k 
& 12k 
& -- 
& 50 
& 1 
& 5 
& Autonomous \\

Articubot
& 42.3k 
& 42.3k 
& 2 
& 1 
& 1 
& 1 
& Autonomous \\

MolmoBot
& 1.7M 
& 125.6k 
& 2 
& 1 
& 1 
& 1 
& Autonomous \\

InternData-A1
& 630k 
& 74.4k 
& 18 
& 70 
& 227
& 4
& Autonomous \\

\midrule

\textbf{\ourdataset}
& \textbf{1M}
& \textbf{1M}
& \textbf{23}
& \textbf{44(Atomic)}
& \textbf{1{,}122}
& \textbf{5}
& \textbf{Autonomous} \\

\bottomrule
\end{tabular}
}
\end{table}

%% file: sections/05_simulation_experiments.tex
\section{Simulation Experiments}
\label{sec:simulation_experiments}

In this section, we evaluate the effectiveness of \ourdataset using two simulation benchmarks: LIBERO~\citep{liu2023libero} and RoboCasa365~\citep{nasiriany2026robocasa365}.

\subsection{Experiment Setup}

We directly adopt the model architecture proposed in PRTS~\citep{zhang2026prts} as the base VLA architecture for our experiments.
The architecture consists of a pretrained Qwen3-VL~\citep{bai2025qwen3} 4B backbone and a diffusion transformer (DiT)-based action expert that generates continuous robot actions via flow matching.
To isolate the effect of pretraining data and enable a controlled comparison with representative VLA models, we retain the architectural design of PRTS while removing its Contrastive Reinforcement Learning (CRL) objectives. 
In the first stage, we train the VLM backbone exclusively on \ourdataset using a standard cross-entropy objective to predict robot action sequences tokenized with the FAST~\citep{pertsch2025fast} tokenizer.
In the second stage, we introduce the DiT-based action expert and train the resulting VLA model on the downstream dataset of each target benchmark using a flow-matching objective. 
Across all controlled variants, we keep the policy architecture and downstream training recipe fixed, such that the primary difference lies in the pretraining data.

We denote the model whose first-stage pretraining uses exclusively \ourdataset as \THEPT. 
For comparison, we reuse the \textit{PRTS w/o CRL} ablation checkpoint evaluated in the PRTS technical report~\citep{zhang2026prts}, which is pretrained on real-world data, and denote it as \THEBASE. 
We additionally train a variant that skips first-stage pretraining and undergoes only second-stage training on the downstream benchmark dataset, denoted as \THEFS. 
See App.~\ref{app:subsec:training_detail} for model and training details.

\subsection{Key Findings}

\begin{table}[htp]
    \centering
    \caption{\textbf{Evaluation results (success rates) on LIBERO.}
    Training budgets are shown in gray parentheses.
    The best result in each column is shown in \textbf{bold}.}
    \label{tab:libero_main}
    \small
    \begin{tabular*}{\linewidth}{@{\extracolsep{\fill}}lccccc@{}}
        \toprule
        Method & Spatial & Object & Goal & Long & Average \\
        \midrule
        GR00T N1 & 94.4 & 97.6 & 93.0 & 90.6 & 93.9 \\
        $\pi_{0}$ \textcolor{gray}{(bs=32, 30K steps)} & 96.8 & 98.8 & 95.8 & 85.2 & 94.2 \\
        Qwen3-VL-$\pi$ \textcolor{gray}{(bs=32, 30K steps)} & 95.2 & 99.0 & 96.2 & 88.4 & 94.7 \\
        InternVLA-M1 & 98.0 & 99.0 & 93.8 & 92.6 & 95.9 \\
        $\pi_{0.5}$ \textcolor{gray}{(bs=256, 30K steps)} & \textbf{98.8} & 98.2 & \textbf{98.0} & 92.4 & 96.9 \\
        \midrule
        \thebase \textcolor{gray}{(bs=32, 30K steps)} & 97.8 & \textbf{99.8} & \textbf{98.0} & \textbf{95.6} & \textbf{97.8} \\
        \thefs \textcolor{gray}{(bs=32, 30K steps)} & 95.8 & 99.2 & 95.8 & 88.0 & 94.7 \\
        \thept \textcolor{gray}{(bs=32, 30K steps)} & 97.6 & 98.8 & 96.8 & 94.0 & 96.8 \\
        \bottomrule
    \end{tabular*}
\end{table}



\begin{table}[htp]
    \centering
    \caption{\textbf{Evaluation results (success rates) on RoboCasa365.}
    Best results are in bold. 
    \thebase is omitted here because the official report does not provide results for the variant without CRL.
    }
    \label{tab:robocasa_main}
    \small
    \begin{tabular*}{\linewidth}{@{\extracolsep{\fill}}lcccc@{}}
        \toprule
        Method & Atomic Seen & Composite Seen & Composite Unseen & Average \\
        \midrule
        $\pi_{0}$ & 34.6 & 6.1 & 1.1 & 14.8 \\
        $\pi_{0.5}$ & 39.6 & 7.1 & 1.2 & 16.9 \\
        GR00T N1.6 & 51.1 & 9.4 & 1.7 & 21.9 \\
        GR00T N1.5 & 50.7 & 14.8 & 2.7 & 23.9 \\
        \midrule
        \thefs & 43.3 & 9.4 & 4.4 & 20.0 \\
        \thept & \textbf{52.8} & \textbf{23.1} & \textbf{7.5} & \textbf{28.8} \\
        \bottomrule
    \end{tabular*}
\end{table}

\textbf{Pretraining on \ourdataset is effective.}
We compare \thept against \thefs on both LIBERO and RoboCasa365 while holding the model architecture and post-training recipe fixed.
On LIBERO (Tab.~\ref{tab:libero_main}), under the standard post-training configuration (bs=32, 30k steps), \thept achieves an average success rate of 96.8\%, outperforming \thefs (94.7\%) by 2.1 points and surpassing $\pi_{0}$ (94.2\%) and Qwen3-VL-$\pi$ (94.7\%).
It essentially matches $\pi_{0.5}$ (96.9\%) despite using a batch size of 32.
On RoboCasa365 (Tab.~\ref{tab:robocasa_main}), which features a substantially different task distribution and more complex tasks than LIBERO, \thept achieves an average score of 28.8\%, compared with 20.0\% for \thefs, and outperforms every compared baseline, including GR00T N1.5 (23.9\%), GR00T N1.6 (21.9\%), $\pi_{0.5}$ (16.9\%), and $\pi_{0}$ (14.8\%).
While these baselines are pretrained on real-world data, with the GR00T series additionally incorporating simulated and generated data, \thept is pretrained exclusively on synthetic data and still remains competitive on both benchmarks.
Our pretraining data also consists predominantly of articulated-object manipulation, which differs substantially from LIBERO's largely pick-and-place task distribution, yet the benefit persists.
Together, these results show that pretraining on \ourdataset provides effective and transferable manipulation priors.

\begin{figure}[htp]
    \centering
    \includegraphics[width=0.8\linewidth]{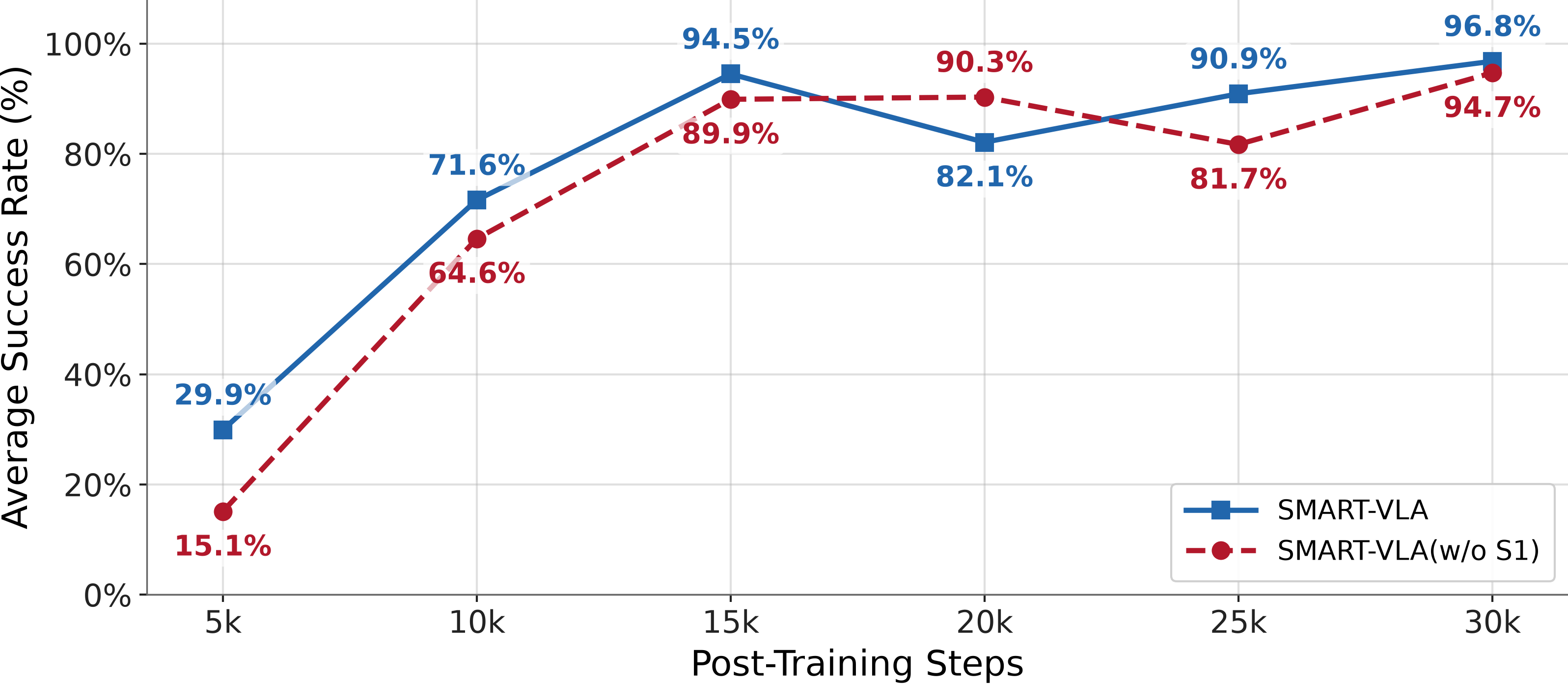}
    \caption{\textbf{Post-training performance curve of \thept and \thefs on LIBERO.}
    Average success rates are reported every 5K post-training steps up to the standard 30K-step training budget.
    }
    \label{fig:libero_convergence}
\end{figure}

Convergence behavior on LIBERO provides further evidence for the effectiveness of \ourdataset (Fig.~\ref{fig:libero_convergence}).
On LIBERO, \thept reaches an average success rate of 29.9\% after 5k post-training steps, compared with 15.1\% for \thefs at the same step count.
At 10k steps, \thept reaches 71.6\%, compared with 64.6\% for \thefs.
By 15k steps, \thept reaches 94.5\%, whereas \thefs reaches 89.9\%.
The \thept curve is non-monotonic after 15k steps: it falls to 82.1\% at 20k steps, recovers to 90.9\% at 25k steps, and reaches 96.8\% at 30k steps.
At 30k steps, \thept achieves a 2.1-point advantage over \thefs (96.8\% vs.\ 94.7\%).
Together, the stronger early performance and higher 30k-step result demonstrate that pretraining on \ourdataset provides useful manipulation priors that facilitate downstream task adaptation.


\textbf{\thept shows superior performance on composite and long-horizon tasks.}
On LIBERO-Long, \thept reaches 94.0\%, compared with 88.0\% for \thefs, 85.2\% for $\pi_{0}$, 88.4\% for Qwen3-VL-$\pi$, 90.6\% for GR00T N1, 92.4\% for $\pi_{0.5}$, and 92.6\% for InternVLA-M1.
Only \thebase, which is pretrained on real-world data, achieves a higher score of 95.6\%.
On RoboCasa365, \thept reaches 52.8\% on Atomic Seen, 23.1\% on Composite Seen, and 7.5\% on Composite Unseen.
The strongest non-SMART baselines score 51.1\%, 14.8\%, and 2.7\% on these splits, respectively, so the relative advantage of \thept grows from 3.3\% on Atomic Seen to 56.1\% on Composite Seen and 177.8\% on Composite Unseen.
Compared with \thefs, \thept improves by 6.0 points on LIBERO-Long, 13.7 points on Composite Seen, and 3.1 points on Composite Unseen.
This pattern is consistent with the multi-stage nature of articulated-object manipulation in \ourdataset and suggests that pretraining on such demonstrations may improve temporal generalization.
These results motivate further study of how the temporal structure of pretraining data affects compositional control.

%% file: sections/06_real_world_experiment.tex
\section{Sim-to-Real Experiments}

In this section, we conduct real-robot experiments on three robot platforms to further evaluate the effectiveness of \ourdataset and our data synthesis pipeline in downstream articulated-object manipulation tasks.
We investigate the following three research questions (RQs):

\begin{itemize}
    \item \textbf{RQ1.} How effective is pretraining at improving downstream performance on articulated-object manipulation tasks, compared to training from scratch? 
    \item \textbf{RQ2.} Is \ourframework capable of generating effective simulation data to produce a feasible policy on real hardware, and does this data-generation pipeline generalize consistently across different robot embodiments? 
    \item \textbf{RQ3.} Can the performance of VLA models on articulated-object manipulation tasks be scaled up with the amount of synthesized data? 
\end{itemize}

\subsection{Experiment Setup}

\begin{figure}[htp]
    \centering
    \includegraphics[width=0.9\linewidth]{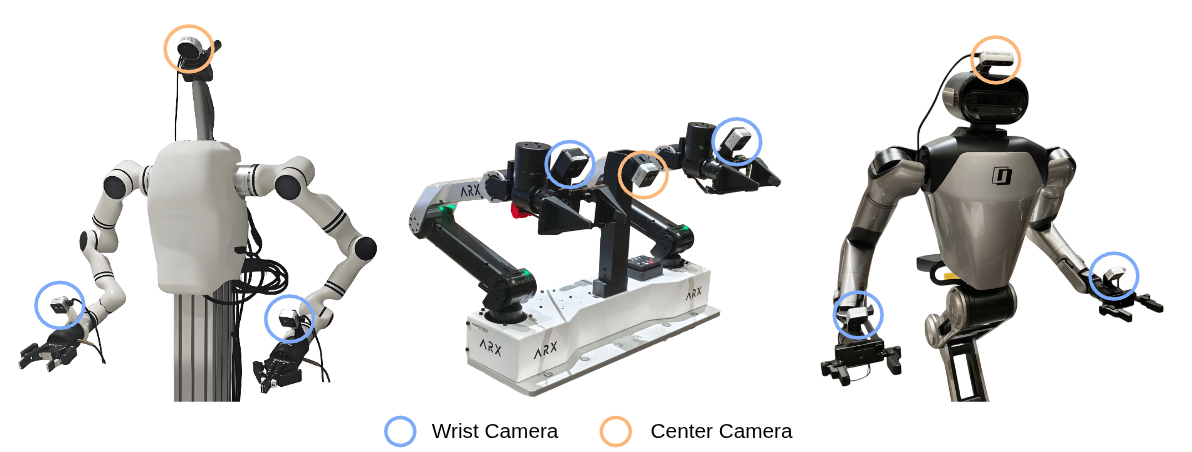}
    \caption{Overview of the real-world robot platforms, including a dual-arm RealMan RM75 platform (left), an AC1 dual-arm platform (center), and an R1Pro platform (right).
    All platforms have two wrist cameras and one center-mounted head camera.}
    \label{fig:rm_dual_arm_platform}
\end{figure}

\textbf{Experiment platform.}
Our real-world evaluations use three platforms: a dual-arm RealMan RM75 platform, an R1Pro platform, and an AC1 platform, as shown in Fig.~\ref{fig:rm_dual_arm_platform}.
All three platforms are dual-arm robots equipped with three RGB cameras, including one center-mounted head camera and one wrist camera on each arm.
The camera intrinsics and extrinsics are aligned between simulation and real-world tasks to keep sim-to-real consistency.

\textbf{Task design and data synthesis.} 
We construct twelve real-world articulated-object manipulation tasks (8 types and 4 cross-platform variations) for evaluation, spanning prismatic, revolute, and compound articulation mechanisms as well as single-stage and multi-stage interactions.
The eight task types are:

\begin{itemize}[leftmargin=*, nosep]
    \item \textbf{Bucket-handle lifting}: lift the bucket by its compliant revolute handle, requiring stable contact under handle deformation.
    \item \textbf{Drawer pulling}: grasp the drawer handle and pull the drawer along its prismatic rail to open the cabinet.
    \item \textbf{Dispenser pressing}: press a spring-loaded dispenser button through its downward stroke to release water from the container.
    \item \textbf{Cabinet-door sliding}: grasp the door handle and slide the right-side cabinet door leftward along its prismatic rail.
    \item \textbf{Toaster-lever pressing}: push the toaster lever from its upper position to the bottom while maintaining continuous gripper contact.
    \item \textbf{Microwave door closing}: push a horizontally hinged microwave door closed through a revolute motion until the latch is engaged.
    \item \textbf{Microwave knob rotation}: grasp the upright microwave knob and rotate it clockwise through a constrained revolute motion.
    \item \textbf{Toolbox-lid flipping}: grasp the upper-lid handle of a horizontally placed toolbox and rotate the lid upward to open the box.
\end{itemize}

\input{figures/real_task_illustration}

Detailed task designs, cross-platform assignments, initial states, and success criteria are provided in App.~\ref{app:subsec:task_details}.

For post-training, we synthesize task-specific demonstrations using our data synthesis pipeline.
To diversify demonstration trajectories and improve policy robustness, we randomize the initial configurations of task objects and robots within small, bounded ranges, and introduce perturbation into motion generation as mentioned in Sec.~\ref{subsec:motion_gen}.
We also randomize scene lighting at each environment reset to diversify the illumination conditions in demonstrations.

\textbf{Training and evaluation protocol.}
We adopt the same model notation as in Sec.~\ref{sec:simulation_experiments}, using only demonstrations generated by \ourframework for the second training phase.
We also post-train a \thebaseline-base model on the same synthesized dataset for comparison.
In the sim-to-real experiment, we train models on the full 500-episode datasets and directly evaluate them on the real-world tasks.
In the performance scaling experiment, we sample 500, 200, 100, 50, 10, and 1 episodes from the full 500-episode datasets recursively to construct subdatasets for post-training.

We evaluate policy performance using two metrics: 
1) success rate (SR), computed as the percentage of successful trials over 15 independent trials for each policy; and 
2) action chunk quality, characterized by normalized root-mean-square joint jerk (NRMSJ) and normalized mean action variation (NMAV).
NRMSJ evaluates local motion smoothness by measuring rapid changes in joint dynamics, while NMAV provides a global assessment of trajectory consistency by computing the mean L2 distance between consecutive normalized action vectors over an action chunk.
See App.~\ref{app:eval_protocol_details} for detailed definitions.

\subsection{Sim-to-Real Experiment Results}

\begin{figure}[htp]
    \centering
    \includegraphics[width=\linewidth]{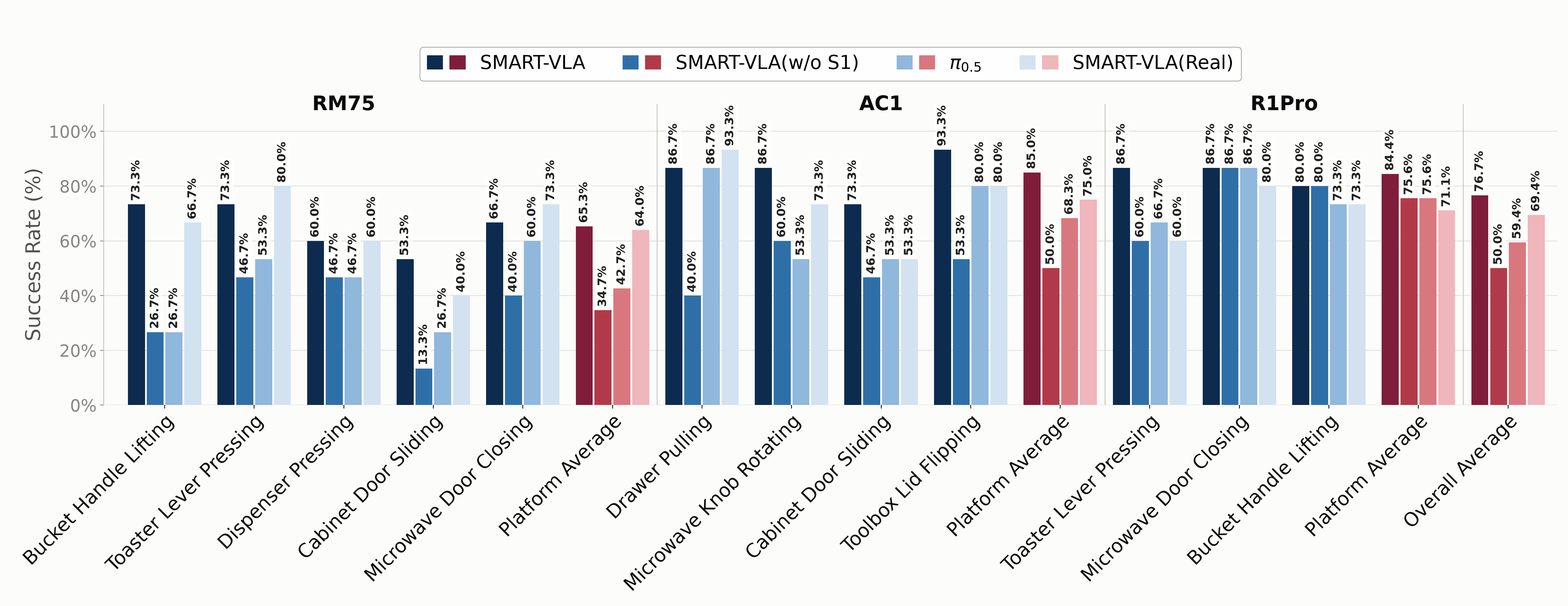}
    \caption{Real-world evaluation across twelve articulated-object manipulation tasks on three robot platforms.}
    \label{fig:real_world_500ep_result}
\end{figure}

We compare four training conditions, \thept, \thefs, \thebase, and \thebaseline, on the twelve real-world articulated-object manipulation tasks across three real-world robot platforms.
The results are shown in Fig.~\ref{fig:real_world_500ep_result}.

\textbf{Demonstrations synthesized by \ourframework can enable zero-shot sim-to-real policy transfer in articulated-object manipulation across different robot platforms.}
Across all evaluated settings, \thept reaches an overall average SR of 76.7\%, peaking at 85.0\% on AC1 and 84.4\% on R1Pro.
It also achieves 93.3\% SR on toolbox-lid flipping with AC1 and 86.7\% on microwave door closing with R1Pro, showing strong performance across different tasks.
This effectiveness holds across policy models, all of which achieve at least 50.0\% overall average success.
Together, these results demonstrate that synthetic data enables effective VLA post-training for articulated-object manipulation and zero-shot sim-to-real transfer across robot platforms, tasks, and policy architectures without real-world demonstrations.
This further demonstrates that \ourframework can generate demonstrations that provide effective supervision for real-world articulated-object manipulation.

\textbf{\thept matches or outperforms comparable policies on real-world articulated-object manipulation tasks across three embodiments, indicating the effectiveness of \ourdataset as a VLA pretraining dataset.}
Across the twelve tasks, \ourdataset pretraining raises SR by 26.7 points on average (76.7\% vs.\ 50.0\%), while \thept matches or outperforms \thefs on every task and attains its largest advantage of 46.6 points on RM75 bucket-handle lifting (73.3\% vs.\ 26.7\%).
This advantage holds across embodiments despite variations in improvement magnitude, suggesting that \ourdataset pretraining provides embodiment-agnostic benefits.
Against the two baselines pretrained on real-world data, \thept attains an average SR of 76.7\%, exceeding $\pi_{0.5}$ and \thebase by 17.3 and 7.3 points and matching or outperforming $\pi_{0.5}$ on ten of the twelve tasks.
Together, these results indicate that \ourdataset pretraining is competitive with real-world pretraining across the evaluated articulated-object manipulation tasks.

\textbf{Pretraining on \ourdataset enables smoother action generation.}
As shown in Tab.~\ref{tab:jerk}, \thept achieves the lowest NRMSJ (0.0581) and NMAV (0.0673), substantially lower than the corresponding values for \thefs (0.1592 and 0.1674), and generates smoother, more stable action chunks.
The same trend appears in NMAV, where \thept, \thebase, $\pi_{0.5}$, and \thefs obtain 0.0673, 0.0721, 0.1625, and 0.1674, respectively.
Consistent with these quantitative results, we observe that \thefs generates visibly jerkier and less stable action chunks than \thept, \thebase, and $\pi_{0.5}$ during real-world execution.
Since these models differ primarily in their pretraining data, this trend can be attributed to the motion generation process in \ourdataset.
Collision-free motion planning and in-contact constrained motion generation explicitly encourage local smoothness and global consistency, whereas the quality of real-world teleoperation trajectories is generally controlled less explicitly.
This result suggests that the trajectories synthesized through collision-free motion planning and in-contact constrained motion generation provide structured motion priors for stable action generation.

\begin{table}[t]
\centering
\caption{The average of Normalized Root-Mean-Square Jerk (NRMSJ, normalized using quantiles) and Normalized Mean Action Variation (NMAV, normalized using quantiles) across AC1, RM75, R1Pro, and all twelve tasks. Lower values (\textcolor{blue}{\(\downarrow\)}) indicate smoother and more consistent trajectories.}
\label{tab:jerk}
\setlength{\tabcolsep}{8pt}
\begin{tabular}{lcccccccc}
\toprule
& \multicolumn{4}{c}{NRMSJ \textcolor{blue}{\(\downarrow\)}}
& \multicolumn{4}{c}{NMAV \textcolor{blue}{\(\downarrow\)}} \\
\cmidrule(lr){2-5}
\cmidrule(lr){6-9}
Method
& RM75
& AC1
& R1Pro
& Overall
& RM75
& AC1
& R1Pro
& Overall \\
\midrule

\thept
& 0.0258
& 0.1006
& 0.0448
& 0.0581
& 0.0416
& 0.1046
& 0.0520
& 0.0673 \\

\thebase
& 0.0456
& 0.1177
& 0.0157
& 0.0637
& 0.0584
& 0.1209
& 0.0253
& 0.0721 \\

\thefs
& 0.0357
& 0.3703
& 0.0424
& 0.1592
& 0.0481
& 0.3757
& 0.0486
& 0.1674 \\

\thebaseline
& 0.1636
& 0.0705
& 0.0225
& 0.0913
& 0.0868
& 0.2445
& 0.1539
& 0.1625 \\

\bottomrule
\end{tabular}
\end{table}

\begin{figure}[t]
\centering
\includegraphics[width=\linewidth]{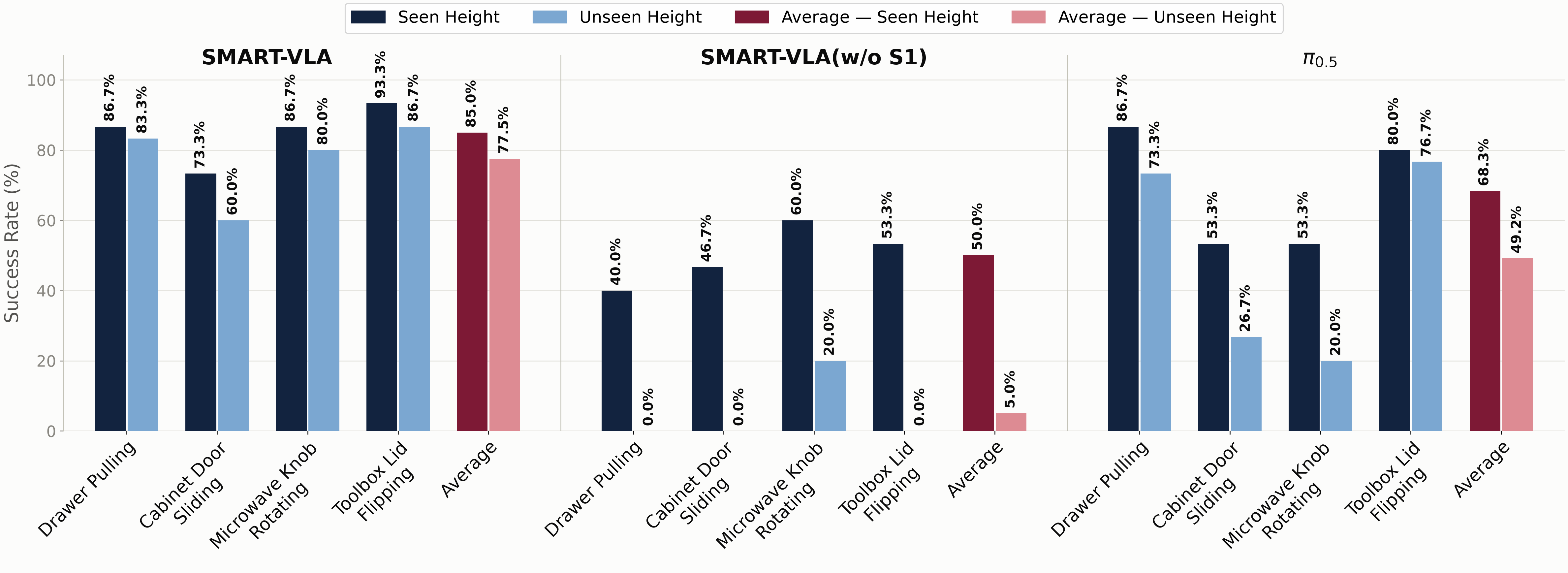}
\caption{The success rates of \thept, \thefs, and \thebaseline on AC1 tasks at seen and unseen environment heights.}
\label{fig:ac1_height}
\end{figure}

\textbf{Pretraining on \ourdataset also improves robustness to height variation.}
As shown in Fig.~\ref{fig:ac1_height}, under an unseen deployment height on AC1 (10 cm higher than during data synthesis), \thept maintains at least 60\% SR across all four tasks with an average drop of only 7.5 points, compared with 19.2 and 45.0 points for $\pi_{0.5}$ and \thefs, respectively.
In contrast, \thefs fails completely on three of the four tasks at this unseen height.
The substantial degradation of \thefs highlights that policies without pretraining are more vulnerable to out-of-distribution deployment conditions.
Despite being pretrained exclusively on synthetic data, \thept exhibits stronger robustness than $\pi_{0.5}$, suggesting that the broader articulation coverage and more diverse manipulation trajectories in \ourdataset provide more transferable manipulation priors.
Together, these results indicate that the base-height variation covered during data synthesis provides sufficient diversity for the policy to transfer to a higher robot base.

\subsection{Performance Scaling Experiment}

\begin{figure}[htp]
    \centering
    \includegraphics[width=0.95\linewidth]{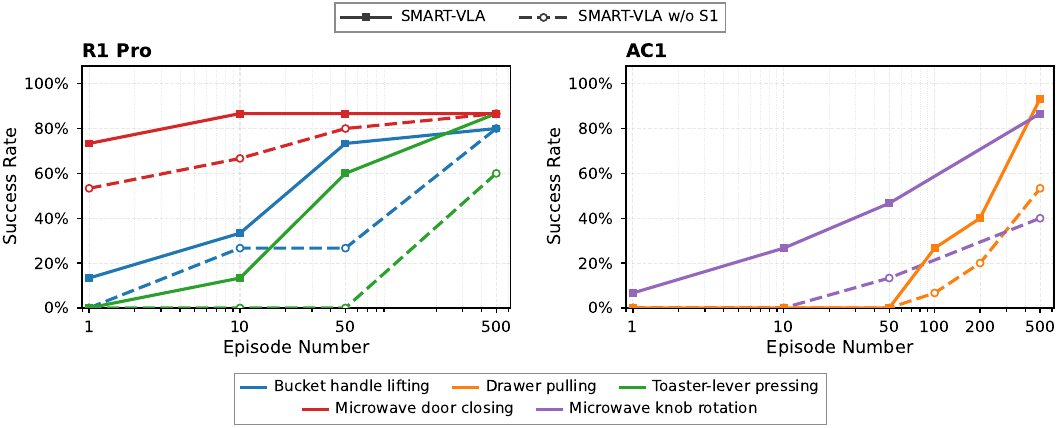}
    \caption{Performance scaling experiment results on AC1 and R1Pro tasks.}
    \label{fig:performance_scaling}
\end{figure}

\textbf{The positive scaling trend suggests that \ourframework provides increasingly rich supervision for articulated-object manipulation.}
Across all five tasks, both \thept and \thefs improve consistently with more synthesized demonstrations.
For \thept, scaling from 1 to 50 episodes increases SR from 13.3\% to 73.3\% on R1Pro bucket-handle lifting, from 0\% to 60.0\% on toaster-lever pressing, and from 6.7\% to 46.7\% on AC1 microwave knob rotation.
More challenging tasks require further scaling for \thept: drawer pulling shows no success below 50 episodes but reaches 40.0\% at 200 episodes, while all five tasks reach at least 80.0\% at 500 episodes.
Scaling rates track task complexity, with loosely constrained tasks (microwave door closing, bucket-handle lifting) largely saturating by 50 episodes, whereas toaster-lever pressing, drawer pulling, and microwave knob rotation keep improving at larger scales.
Together with spatial and lighting randomization, these results show that \ourframework provides scalable, diverse supervision, with larger synthetic datasets most beneficial for tasks requiring precise contacts and constrained trajectories.

\textbf{Pretraining on \ourdataset enables more effective and data-efficient downstream adaptation for sim-to-real policy transfer.}
As shown in Fig.~\ref{fig:performance_scaling}, \thept matches or outperforms \thefs at the same post-training data budget.
With just one post-training episode, \thept already shows non-trivial manipulation capabilities, reaching 73.3\%, 13.3\%, and 6.7\% SR on microwave door closing, bucket-handle lifting, and microwave knob rotation.
Pretraining also cuts the task-specific data needed for strong performance, as 50 episodes bring 73.3\% on R1Pro bucket-handle lifting and 60.0\% on toaster-lever pressing, close to \thefs at 500 episodes (80.0\% and 60.0\%).
Pretraining benefits are particularly clear on microwave knob rotation, where \thefs records no success with up to 10 episodes while \thept succeeds with one.
Overall, these results show that \ourdataset pretraining enables rapid adaptation from limited task-specific data while maintaining consistent gains as downstream data scales.

%% file: figures/real_task_illustration.tex
\begin{figure}[t]
    \centering
    \begin{subfigure}[t]{0.235\linewidth}
        \centering
        \includegraphics[width=\linewidth]{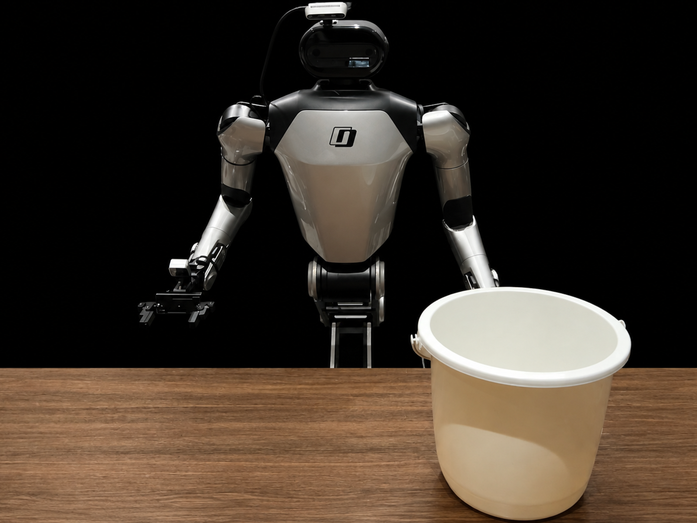}
        \caption{R1Pro bucket-handle lifting.}
        \label{fig:task_bucket_handle_1}
    \end{subfigure}
    \hfill
    \begin{subfigure}[t]{0.235\linewidth}
        \centering
        \includegraphics[width=\linewidth]{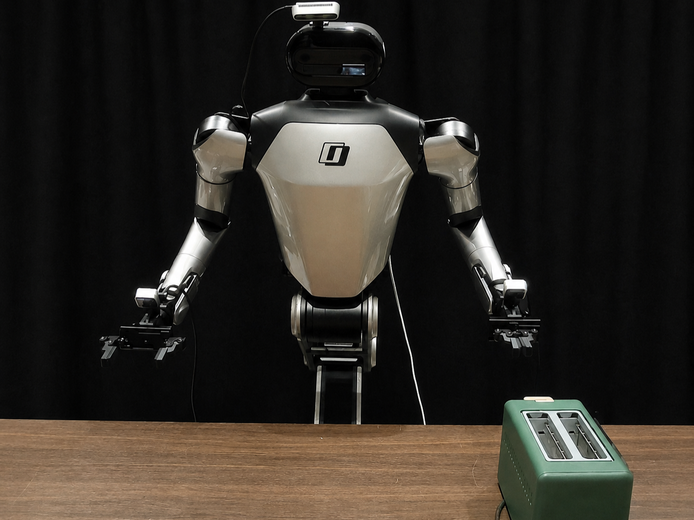}
        \caption{R1Pro toaster-lever pressing.}
        \label{fig:task_bucket_handle_2}
    \end{subfigure}
    \hfill
    \begin{subfigure}[t]{0.235\linewidth}
        \centering
        \includegraphics[width=\linewidth]{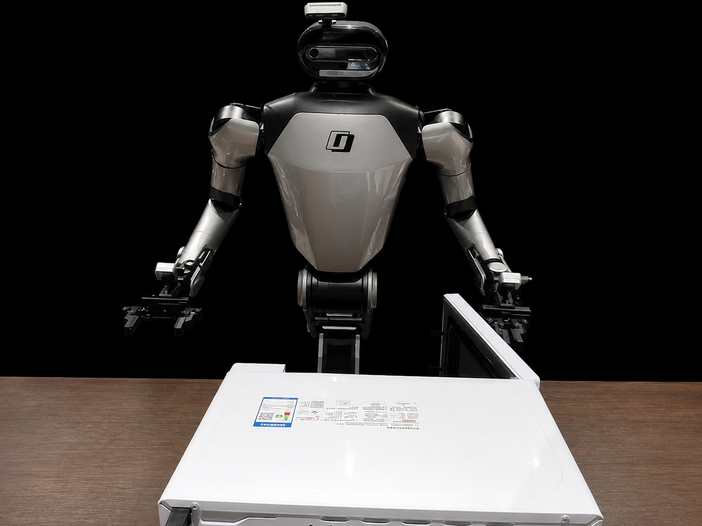}
        \caption{R1Pro microwave door closing.}
        \label{fig:task_drawer_opening_1}
    \end{subfigure}
    \hfill
    \begin{subfigure}[t]{0.235\linewidth}
        \centering
        \includegraphics[width=\linewidth]{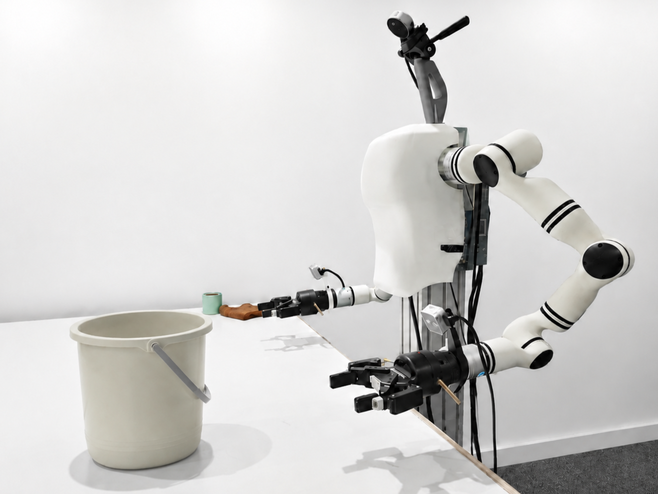}
        \caption{RM75 bucket-handle lifting.}
        \label{fig:task_drawer_opening_2}
    \end{subfigure}

    \medskip

    \begin{subfigure}[t]{0.235\linewidth}
        \centering
        \includegraphics[width=\linewidth]{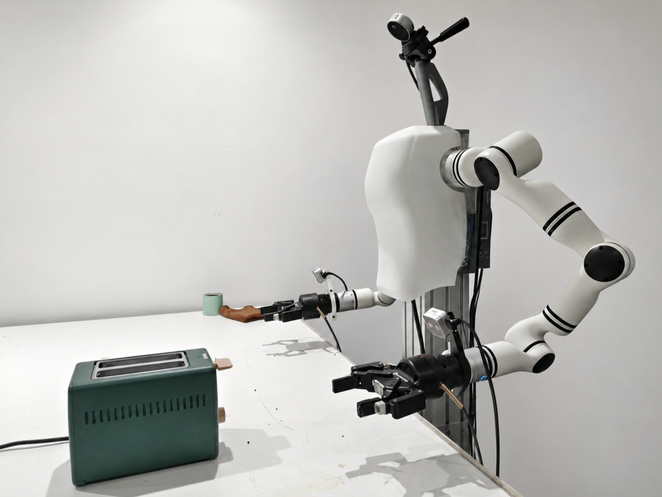}
        \caption{RM75 toaster-lever pressing.}
        \label{fig:task_dispenser_pressing}
    \end{subfigure}
    \hfill
    \begin{subfigure}[t]{0.235\linewidth}
        \centering
        \includegraphics[width=\linewidth]{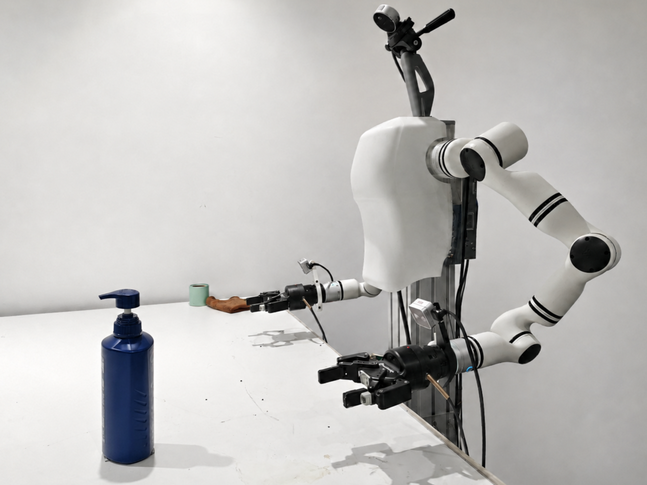}
        \caption{RM75 dispenser pressing.}
        \label{fig:task_sliding_door_1}
    \end{subfigure}
    \hfill
    \begin{subfigure}[t]{0.235\linewidth}
        \centering
        \includegraphics[width=\linewidth]{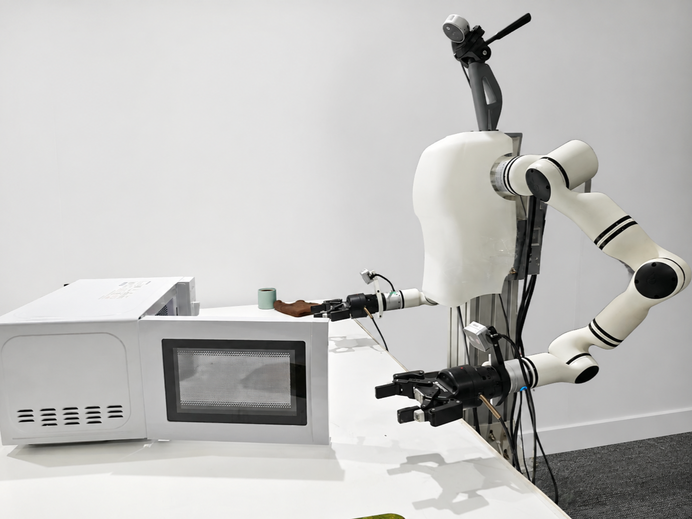}
        \caption{RM75 microwave door closing.}
        \label{fig:task_sliding_door_2}
    \end{subfigure}
    \hfill
    \begin{subfigure}[t]{0.235\linewidth}
        \centering
        \includegraphics[width=\linewidth]{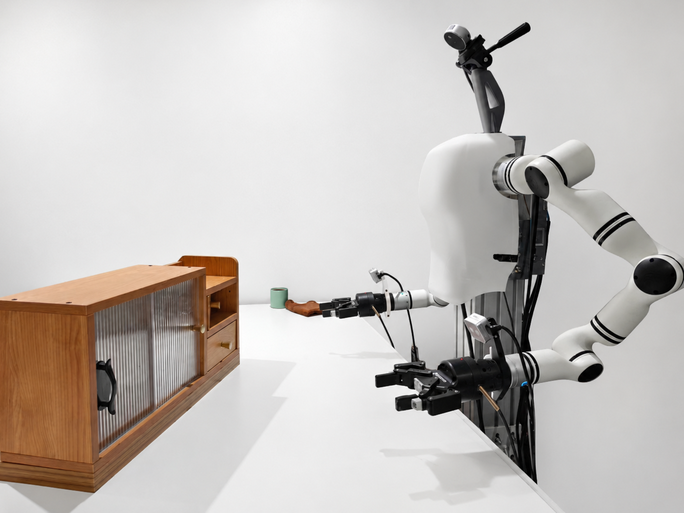}
        \caption{RM75 cabinet-door sliding.}
        \label{fig:task_toaster_lever}
    \end{subfigure}

    \medskip

    \begin{subfigure}[t]{0.235\linewidth}
        \centering
        \includegraphics[width=\linewidth]{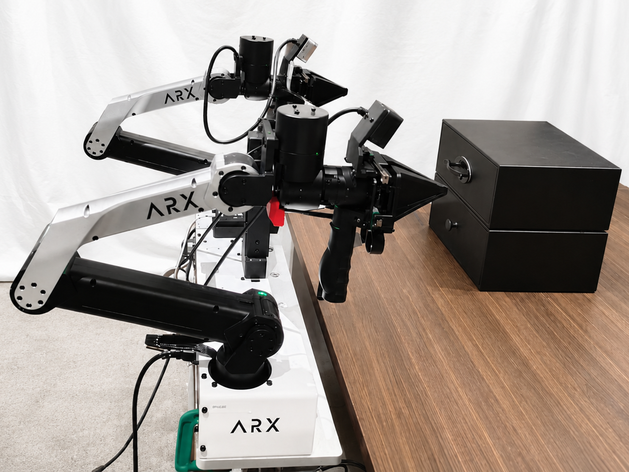}
        \caption{AC1 drawer pulling.}
        \label{fig:task_microwave_door}
    \end{subfigure}
    \hfill
    \begin{subfigure}[t]{0.235\linewidth}
        \centering
        \includegraphics[width=\linewidth]{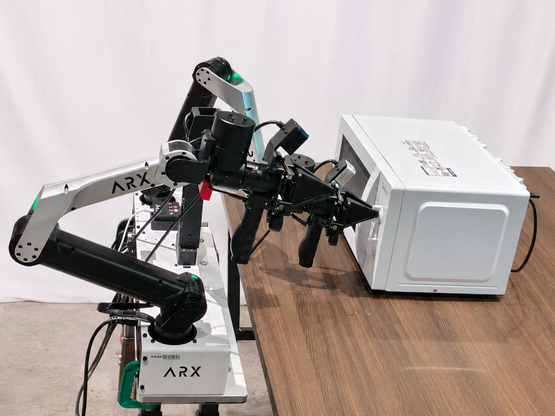}
        \caption{AC1 microwave knob rotation.}
        \label{fig:task_microwave_knob}
    \end{subfigure}
    \hfill
    \begin{subfigure}[t]{0.235\linewidth}
        \centering
        \includegraphics[width=\linewidth]{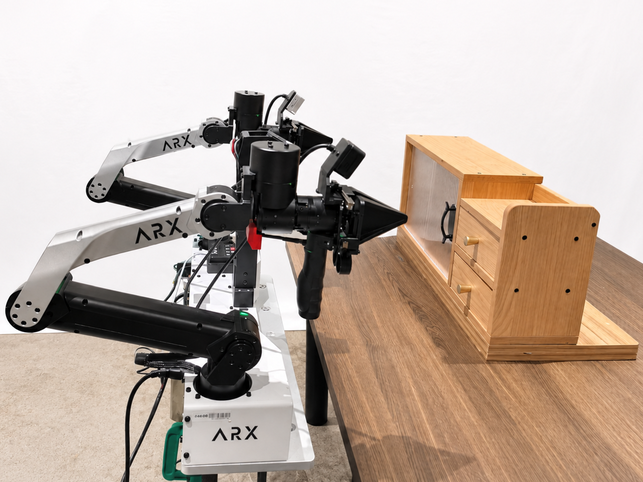}
        \caption{AC1 cabinet-door sliding.}
        \label{fig:task_microwave_sequence}
    \end{subfigure}
    \hfill
    \begin{subfigure}[t]{0.235\linewidth}
        \centering
        \includegraphics[width=\linewidth]{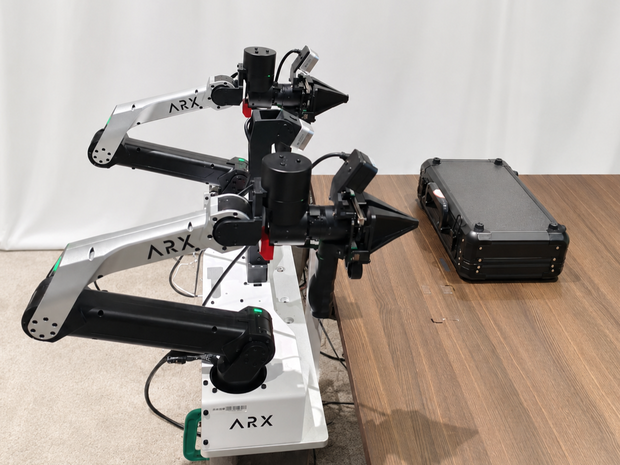}
        \caption{AC1 toolbox-lid flipping.}
        \label{fig:task_toolbox_lid}
    \end{subfigure}

    \caption{Overview of the twelve real-world articulated-object manipulation tasks.}
    \label{fig:real_world_tasks}
\end{figure}

%% file: sections/07_discussion.tex
\section{Limitations}

Our work has three limitations.
First, synthesis throughput decreases for long-horizon tasks due to lower planning and execution success rates.
One solution is to synthesize shorter subtasks independently and reconnect them through consistent boundary states.
Second, agentic task generation is susceptible to LLM hallucinations and depends on the underlying LLM/VLM capabilities.
Stronger verification and human-in-the-loop checks can reduce such errors.
Third, articulated assets require substantial post-processing for geometry separation and articulation annotation.
Future work could jointly generate articulation attributes and manipulation annotations to reduce manual effort.

\section{Conclusion}

In this work, we present \ours, a scalable system for synthesizing articulated-object manipulation demonstrations for VLA training.
At its core, \ourframework supports automatic motion generation, diverse robot embodiments, and extensive domain randomization.
Building on \ourframework, agentic task generation and our distributed data synthesis system enable scalable construction of \ourdataset, covering diverse robot setups, articulated objects, and manipulation skills.
Experiments on simulation benchmarks and real-world tasks show that pretraining on \ourdataset improves downstream VLA performance and data efficiency and enables zero-shot sim-to-real transfer.
Performance also scales positively with the number of task-specific synthetic demonstrations used for post-training.
Together, these results demonstrate that \ours can generate high-quality synthetic demonstrations and suggest that synthetic data has potential as a complement to real-world demonstrations for articulated-object manipulation.

In the real world, collecting large-scale demonstrations for articulated-object manipulation remains challenging due to the difficulty of simultaneously achieving high-quality physical interactions, efficient data acquisition, and broad task diversity. 
Our work shows that simulation provides a promising complementary approach by enabling precise modeling of object articulation and contact-rich interaction processes while maintaining scalability in data generation. 
By bridging scalable synthetic experience generation with VLA learning, we envision synthetic data as an important complementary source that can reduce reliance on costly real-world demonstrations and scale manipulation experience across diverse tasks, objects, and robot embodiments.

%% file: sections/08_contributions.tex
\section*{Contributions}

The contributions of each author are summarized below:

\textbf{Jicong Ao} led the project, designed the overall architecture of the simulation framework, deployed and optimized large-scale data synthesis on the compute cluster, designed and constructed simulation tasks for sim-to-real evaluation, and conducted sim-to-real experiments on the R1Pro and RM75 platforms.

\textbf{Shuhan Jiang} was a core developer of the simulation framework, implemented the distributed data synthesis pipeline, conducted experiments on simulation benchmarks, and conducted sim-to-real experiments on the AC1 platform.

\textbf{Yuling Zhong} contributed to the development of several framework features, conducted data synthesis for the RM75 and R1Pro platforms, was responsible for the sim-to-real experiments on the RM75 platform, and contributed to experiments on the R1Pro platform.

\textbf{Yanwen Liu} developed the domain randomization modules of the simulation framework, resolved critical technical issues to ensure its reliable operation, conducted the initial data synthesis trials, and performed preliminary post-training experiments with synthesized data.

\textbf{Yuhan Gao} provided SDGScenes-generated assets as background scenes for task construction and conducted large-scale articulated-object annotation and task configuration generation for data synthesis on the compute cluster.

\textbf{Jiangyuan Zhao, Yang Zhang} provided PRTS pretraining infrastructure, conducted model pretraining, and delivered proficient analysis in post-training.

\textbf{Chenjia Bai} advised and supervised the project.

\section*{Acknowledgement}

This project was made possible by the support and contributions of many colleagues at TeleAI. 
We would like to thank Wentao Ma for exploring the effective annotation form for articulated-object manipulation, Chenyu Hui for developing the primary annotation pipelines for rigid and articulated objects and organizing the task configuration files, 
Jingyi Deng for contributing to the development of the simulation platform prototype and the design of the manipulation task prototypes, 
Yiqian Xu for providing analysis to optimize the data synthesis pipeline on the compute cluster,
and Ouyang Lu for providing VLA deployment toolkits used in the sim-to-real experiments.

%% file: sections/99_appendix.tex
\appendix

\section{Framework Details}

\subsection{Task File Illustration}
\label{app:subsec:task_file}

The following YAML file shows the task file skeleton used for the toolbox opening task.

\begin{mycodebox}{Task file skeleton for toolbox opening task on AC1 robot}
task:
  task_name: toolbox_open_real_ac1_left_000
  task_type: revolute_door_open_horizontally
  task_description: AC1 robot, use the left arm to open the toolbox.
  detailed_task_description: The AC1 robot first uses the left arm to reach the handle of the toolbox upper lid. Then it opens the toolbox with locked-orientation revolute rotation, flipping the lid upward.
  
data_saving:
  DATA_SAVING_TASK_DIR: toolbox_open_real_ac1_left_000
  trajectory_collecting:
    trajectory_collecting_saving_sub_dir: trajectories
    saving_workers: 4
    task_execution_number: 1000
  replay_demo:
    replay_demo_consuming_sub_dir: trajectories
    replay_demo_saving_sub_dir: demos
    saving_workers: 4
    task_execution_number: 9999
    
simulation:
  trajectory_collecting_num_envs: 100
  trajectory_collecting_time_sync_mode: E_mode
  trajectory_collecting_record_freq: 30
  replay_demo_num_envs: 16
  replay_demo_record_freq: 30
  replay_demo_time_sync_mode: E_mode
  env_spacing: 50
  enable_cameras: true
  render:
    rendering_mode: quality
    rtx.rendermode: RealTimePathTracing
  physics_material:
    static_friction: 0.2
    dynamic_friction: 0.2
    
lighting:
  enable_light_randomization_at_reset: true
  enable_stepwise_light_randomization: false
  global_dome_light:
    color: [..., ...]
    color_temperature: [..., ...]
    intensity: [..., ...]
    texture_format: automatic
    texture_file: ...
  local:
    key_fill_sphere:
      type: sphere
      pos: ...
      color: [..., ...]
      color_temperature: [..., ...]
      intensity: [..., ...]
      radius: [..., ...]
    rim_light_cylinder:
      type: cylinder
      pos: ...
      rot: ...
      color: [..., ...]
      color_temperature: [..., ...]
      intensity: [..., ...]
      radius: ...
      length: ...
      
background:
  use_default_skybox: false
  
robot:
  trajectory_collecting_setup: ac1_cfg_no_camera
  replay_demo_setup: ac1_cfg_camera_no_randomization
  origin_pose:
    translation: [0.05, 0, 0.9]
    eulerzyx: [0, 0, 0]

rigid_objects:
  table_brown:
    usd_path: null
    pos: [0.6074, 0, 0.6585]
    rot: [0, 0.7071, 0.7071, 0]
    scale: [2.0000, 1.8000, 1.0000]
    is_static: true
    pose_randomization_cfg: null
    mdl_material_randomization_cfg: null
    enable_sensors: false
    filter_prim_paths: null
    collision_enabled: false

  scene_lab:
    as_name: scene_lab
    usd_path: null
    pos: [-1.3889, 1.2404, 1.3347]
    rot: [0.9970, -0.0072, 0.0177, 0.0745]
    scale: [1.0, 1.0, 1.0]
    is_static: true
    pose_randomization_cfg: null
    mdl_material_randomization_cfg: null
    enable_sensors: false
    filter_prim_paths: null
    collision_enabled: false

articulated_objects:
  toolbox_real:
    as_name: toolbox_real
    usd_path: null
    is_static: true
    pos: [0.7035, 0.2666, 0.9400]
    rot: [0, -0.7071, 0.7071, 0]
    scale: [1.0, 1.0, 1.0]
    enable_self_collisions: false
    pose_randomization_cfg:
      mode: delta
      position:
        x:
          dist: uniform
          low: -0.05
          high: 0.05
        y:
          dist: uniform
          low: -0.05
          high: 0.05
        z:
          dist: uniform
          low: -0.03
          high: 0.03
      orientation:
        yaw:
          dist: uniform
          low: -0.5
          high: 0.5
    joint_initial_state_dict: null
    disable_gravity: true
    enable_sensors: false
    mdl_material_randomization_cfg: null

task_condition:
  runtime_checking_interval: 0
  duplicate_runtime_conditions_to_ending: false
  conditions:
    all_actor_ee_translation_deviation_condition:
      phase: runtime
      type: failure
      threshold: 0.05
    all_actor_joint_jerk_condition:
      phase: runtime
      type: failure
      threshold: 1
    all_actor_joint_out_of_hard_limit_condition:
      phase: runtime
      type: failure
      threshold: 3.14
    object_fully_opened_condition:
      object_name: toolbox_real
      joint_name: joint_revolute_000
      target_openness: 0.6
      threshold: 0.2
      phase: ending
      type: success
      
actions:
  parallel_rotate_revolute_door_by_angle_with_locked_orientation:
    target_object: toolbox_real
    side: left
    action_frame_name: grasp_001
    robot_action_frame_name: grasp_001
    rotate_angle: 65.0
    approaching_random_offset_params:
      translation_min: [-0.03, -0.02, -0.05]
      translation_max: [0.03, 0.0, -0.03]
      eulerzyx_min: [-5, 0, -5]
      eulerzyx_max: [5, 0, 5]
    approaching_method: motion_planning
    reaching_random_offset_params: null
    reaching_method: interpolation
    grasping_hand_command_name: grasp
    rotate_frame_suffix: revolute_000
    releasing_hand_command_name: grasp
    target_grasp_percentage: null
    retreating_random_offset_params:
      translation_min: [0, 0, -0.05]
      translation_max: [0, 0, -0.02]
      eulerzyx_min: [-5, -10, -5]
      eulerzyx_max: [5, 0, 5]
    retreating_method: skip
\end{mycodebox}

\subsection{Asset Annotation}
\label{app:subsec:asset_annotation}
\begin{figure}[H]
    \centering

    \begin{subfigure}[t]{0.9\linewidth}
        \centering
        \includegraphics[width=\linewidth]{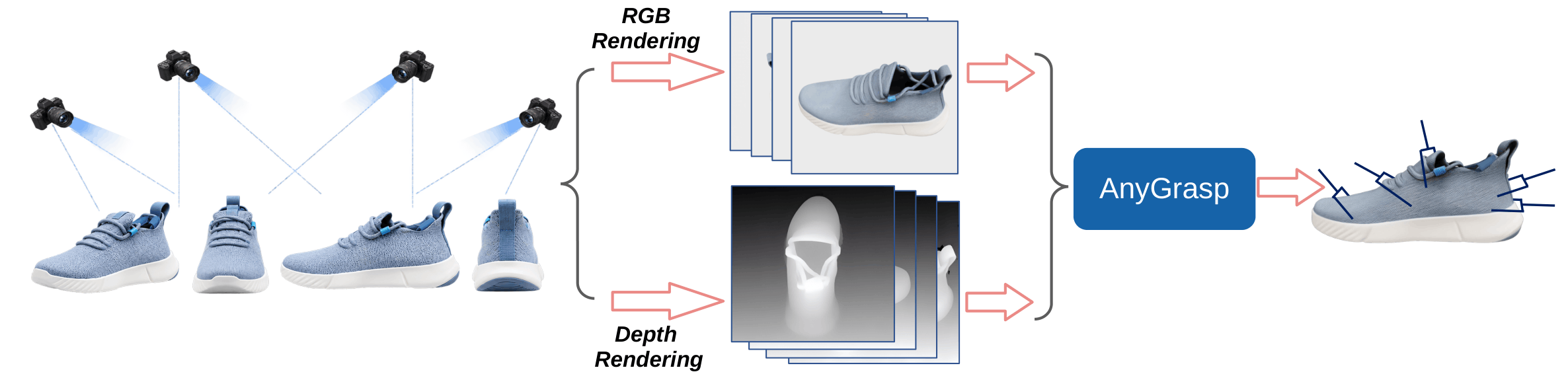}
        \caption{The annotation approach for rigid objects.}
        \label{app:subfig:rigid_anno}
    \end{subfigure}

    \begin{subfigure}[t]{0.9\linewidth}
        \centering
        \includegraphics[width=\linewidth]{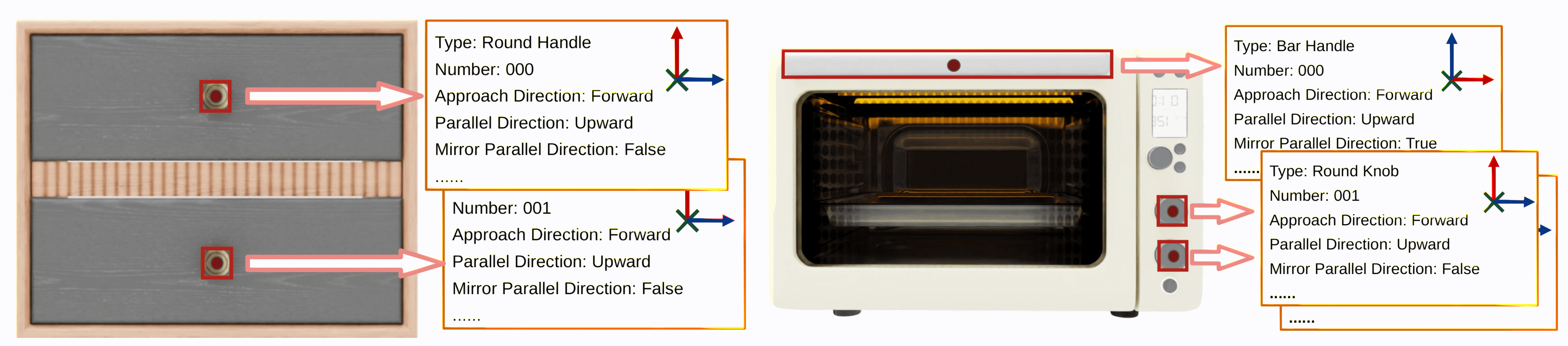}
        \caption{The annotation approach for articulated objects.}
        \label{app:subfig:arti_anno}
    \end{subfigure}

    \caption{The illustration of annotation approaches for (a) rigid objects and (b) articulated objects.}
    \label{fig:app:annotation}
\end{figure}

We illustrate the grasping pose annotation processing introduced in Sec.~\ref{subsec:manipulation_annotation} in Fig.~\ref{fig:app:annotation}.

\subsection{Constraint-Aware Motion Solver}
\label{app:subsec:constraint_solver}

This subsection provides the solver-level details of the constraint-aware motion generation introduced in Sec.~\ref{subsec:motion_gen}.
Given a long-horizon task, the task generation agent produces a grounded sequence of manipulation skills and arm movements.
Each skill specifies the manipulated part, an annotated action frame, and continuous parameters such as a translation distance or a joint rotation angle.
\ourframework then converts each skill into a constrained motion problem.

We represent a movement process by $T+1$ discrete time steps.
Let $t\in\{0,1,\ldots,T\}$ index these steps, where $t=0$ and $t=T$ denote the initial and final steps, respectively.
At time step $t$, the robot joint configuration is $\theta_t\in\mathbb{R}^{n}$, and the pose of the end-effector action frame is ${}^W\mathbf{e}_t\in SE(3)$.
The end-effector pose is determined by forward kinematics as ${}^W\mathbf{e}_t=f_{\mathrm{FK}}(\theta_t)$.
We use $c_j$ to denote a constraint on the robot joint configuration and $c_e$ to denote a constraint on the end-effector pose.
For each skill, let $\mathcal{C}_{\mathrm{goal}}$ contain the end-effector pose constraints $c_e$ that must be satisfied at the target configuration, and let $\mathcal{C}_{\mathrm{path}}$ contain the joint constraints $c_j$ that must be satisfied throughout the movement.

\textbf{Goal configuration.}
The first step solves for the target Cartesian end-effector pose ${}^W\mathbf{e}_T^{\mathrm{tar}}$.
Let $\Delta$ denote the specified motion value of the manipulated joint, such as a displacement for a prismatic joint or a rotation angle for a revolute joint.
The manipulated part annotation provides the action frame ${}^{\mathcal{J}}\mathbf{F}_{\mathrm{act}}$ in the joint coordinate frame.
Based on the joint type and $\Delta$, we construct the corresponding spatial transformation $\mathbf{T}_{\mathrm{joint}}(\Delta)$ in the joint frame.
The canonical target end-effector pose in the joint coordinate frame is obtained by left-multiplying the action frame with this transformation:
\begin{equation}
{}^{\mathcal{J}}\bar{\mathbf{e}}_T
=
\mathbf{T}_{\mathrm{joint}}(\Delta)
\,{}^{\mathcal{J}}\mathbf{F}_{\mathrm{act}}.
\label{app:eq:canonical_target_pose}
\end{equation}
We then express this pose in the global coordinate frame using the joint frame pose ${}^W\mathbf{F}_{\mathcal{J}}$:
\begin{equation}
{}^W\bar{\mathbf{e}}_T
=
{}^W\mathbf{F}_{\mathcal{J}}
\,{}^{\mathcal{J}}\bar{\mathbf{e}}_T.
\label{app:eq:canonical_target_pose_global}
\end{equation}
We use ${}^W\bar{\mathbf{e}}_T$ as the initial target pose for the optimization below and solve for the target end-effector pose that satisfies the skill constraints:
\begin{equation}
\begin{split}
{}^W\mathbf{e}_T^{\mathrm{tar},*} = \arg\min_{{}^W\mathbf{e}_T^{\mathrm{tar}}}\quad
& w_{\mathrm{init}}d_{SE(3)}({}^W\mathbf{e}_T^{\mathrm{tar}},{}^W\bar{\mathbf{e}}_T)^2
  \\
& + \sum_{c_e\in\mathcal{C}_{\mathrm{goal}}}
    w_{\mathrm{pose}}\operatorname{dist}\!\left(c_e({}^W\mathbf{e}_T^{\mathrm{tar}},\Delta),[c_e^{\mathrm{lower}},c_e^{\mathrm{upper}}]\right)^2 \\
&\quad \mathrm{s.t.}\quad
  c_e({}^W\mathbf{e}_T^{\mathrm{tar}},\Delta)\in[c_e^{\mathrm{lower}},c_e^{\mathrm{upper}}],
  \quad \forall\,c_e\in\mathcal{C}_{\mathrm{goal}},
\end{split}
\label{app:eq:goal_pose_solver}
\end{equation}
where $d_{SE(3)}(\cdot,\cdot)$ measures the weighted translation and orientation difference between two end-effector poses, $w_{\mathrm{init}}$ balances the initialization term, and $w_{\mathrm{pose}}$ weights the end-effector pose constraints.

After obtaining ${}^W\mathbf{e}_T^{\mathrm{tar},*}$, we solve for the target joint configuration using a constrained pose-matching objective:
\begin{equation}
\begin{aligned}
\theta_T^{*} = \arg\min_{\theta_T}\quad
& w_{\mathrm{ik}}d_{SE(3)}(f_{\mathrm{FK}}(\theta_T),{}^W\mathbf{e}_T^{\mathrm{tar},*})^2
  + w_{\mathrm{reg}}\lVert\theta_T-\theta_{\mathrm{nominal}}\rVert_2^2
  - w_{\mathrm{manip}}\mathcal{M}(\theta_T) \\
&\quad \mathrm{s.t.}\quad
  \theta_{\min}\leq\theta_T\leq\theta_{\max}, \\
&\phantom{\quad \mathrm{s.t.}\quad}
  \theta_T\in\Theta_{\mathrm{collision\text{-}free}},
\end{aligned}
\label{app:eq:goal_ik}
\end{equation}
where $\mathcal{M}(\theta_T)$ is the manipulability evaluated from the robot Jacobian at $\theta_T$ during the optimization.
This term prefers well-conditioned arm configurations while the joint configuration is being optimized, which is useful for sustained contact tasks where small changes in the end-effector pose can otherwise lead to large joint motions.

\textbf{Constrained trajectory.}
After obtaining $\theta_T^{*}$, the second step solves for the intermediate configurations $\{\theta_t\}_{t=1}^{T-1}$ that connect the initial configuration to the target configuration while preserving the interaction constraints.
We use a constrained trajectory solver of the form
\begin{equation}
\begin{split}
\min_{\{\theta_t\}_{t=1}^{T-1}}\quad
& \sum_{t=1}^{T-1}
  \left(
  w_{\mathrm{s}}\lVert\theta_{t+1}-2\theta_t+\theta_{t-1}\rVert_2^2
  + w_{\mathrm{e}}\lVert\theta_{t+1}-\theta_t\rVert_2^2
  \right) \\
& + \sum_{t=1}^{T-1} w_{\mathrm{c}}
    \sum_{c_j\in\mathcal{C}_{\mathrm{path}}}
    \operatorname{dist}\!\left(c_j(\theta_t),[c_j^{\mathrm{lower}},c_j^{\mathrm{upper}}]\right)^2 \\
&\quad \mathrm{s.t.}\quad
  {}^W\mathbf{e}_t=f_{\mathrm{FK}}(\theta_t), \\
&\phantom{\quad \mathrm{s.t.}\quad}
  c_j(\theta_t)\in[c_j^{\mathrm{lower}},c_j^{\mathrm{upper}}],
  \quad \forall\,c_j\in\mathcal{C}_{\mathrm{path}},\ \forall\,t\in\{1,\ldots,T-1\}, \\
&\phantom{\quad \mathrm{s.t.}\quad}
  \theta_t\in\Theta_{\mathrm{collision\text{-}free}},
  \quad \forall\,t\in\{1,\ldots,T-1\}, \\
&\phantom{\quad \mathrm{s.t.}\quad}
  \theta_0=\theta_{\mathrm{start}},\qquad \theta_T=\theta_T^{*}, \\
&\phantom{\quad \mathrm{s.t.}\quad}
  \lVert\theta_{t+1}-\theta_t\rVert_\infty\leq v_{\max},
  \qquad
  \lVert\theta_{t+1}-2\theta_t+\theta_{t-1}\rVert_\infty\leq a_{\max},
\end{split}
\label{app:eq:path_solver}
\end{equation}
where the three objective terms encourage smooth acceleration, efficient motion, and satisfaction of the path constraints, respectively.
The path constraints $\mathcal{C}_{\mathrm{path}}$ are skill-specific joint constraints derived from the action frame and articulation model, and they restrict the robot joint trajectory to the feasible set induced by the interaction.
The velocity and acceleration bounds are applied to adjacent and consecutive differences of $\theta_t$.

We use the constrained trajectory formulation in Eq.~\eqref{app:eq:path_solver} for in-contact motion, where the articulation model supplies the joint axis, origin, type, and motion-range constraints directly from the simulator.
The resulting joint-space trajectory is checked for collision, joint-limit violation, contact loss, and excessive configuration deviation. Trajectories that violate these conditions are rejected.
This two-stage design allows the same solver interface to support translation-based, rotation-based, and composite manipulation skills while retaining the geometric semantics of the annotated action frames.

\subsection{Randomization Details}
\label{app:subsec:randomization_details}

Tab.~\ref{tab:app:domain_randomization} summarizes the randomization parameters and their sampled ranges.

\begin{table}[H]
\centering
\scriptsize
\setlength{\tabcolsep}{3pt}
\renewcommand{\arraystretch}{1.05}

\caption{Domain randomization parameters of \ourframework.}
\label{tab:app:domain_randomization}

\begin{tabularx}{\linewidth}{@{}>{\raggedright\arraybackslash}p{0.09\linewidth}>{\raggedright\arraybackslash}p{0.20\linewidth}>{\raggedright\arraybackslash}p{0.24\linewidth}>{\raggedright\arraybackslash}X@{}}
\toprule
\textbf{Aspect} & \textbf{Parameter} & \textbf{Range} & \textbf{Purpose} \\
\midrule
Spatial & Object position ($x$, $y$, $z$) & $\pm 0.1$\,m ($x$, $y$), $\pm 0.03$\,m ($z$) & Prevent the policy from memorizing a fixed placement and generalize to arbitrary object positions on the tabletop \\
& Object orientation (roll, pitch, yaw) & $\pm 15^\circ$ (roll, pitch), $\pm 5^\circ$ (yaw) & Cover varied grasp and contact directions so that the policy does not overfit a single pose \\
& Robot initial joint configuration & $\pm 5^\circ$ & Make the policy robust to initial configuration offsets and narrow the gap to real deployment \\
& Camera extrinsics & $\pm 0.05$\,m (position), $\pm 5^\circ$ (orientation) & Tolerate camera mounting errors and ease visual sim-to-real transfer \\
\addlinespace
Physical & Part friction & $\pm 0.15$ & Adapt to contact and slipping behavior under different surface conditions \\
& Part density & $\pm 0.5$ & Cover the dynamics difference caused by part mass variation \\
& Joint stiffness & $\pm 0.003$ (prismatic), $\pm 0.001$ (revolute) & Cover the looseness difference between hinges and sliding rails \\
& Joint damping & $\pm 0.003$ (prismatic), $\pm 0.001$ (revolute) & Cover joint resistance variation and keep motion stable after contact \\
\addlinespace
Visual & Part material and texture & Per part, from a library of 200+ materials and 1000+ textures & Avoid letting the policy rely on a fixed appearance \\
& Metallic & $\pm 0.2$ & Cover the reflectance variation of metal-like surfaces \\
& Roughness & $\pm 0.2$ & Cover gloss and highlight changes under different finishes \\
\addlinespace
Illumination & Light position & $\pm 0.1$\,m ($x$, $y$), $\pm 0.03$\,m ($z$) & Adapt to the shadow and highlight patterns produced by shifted light sources \\
& Light intensity & $[0.5, 1.5]\times$ nominal & Adapt to brightness variation and avoid failures under over- or under-exposure \\
& Light color & $\pm 0.3$ (RGB) & Cover the color casts that appear in the visual input \\
& Light temperature & $\pm 2000$\,K & Cover both warm and cool lighting conditions \\
\bottomrule
\end{tabularx}
\end{table}

\section{Dataset Illustration}
\begin{figure}[H]
\centering
\includegraphics[width=\linewidth]{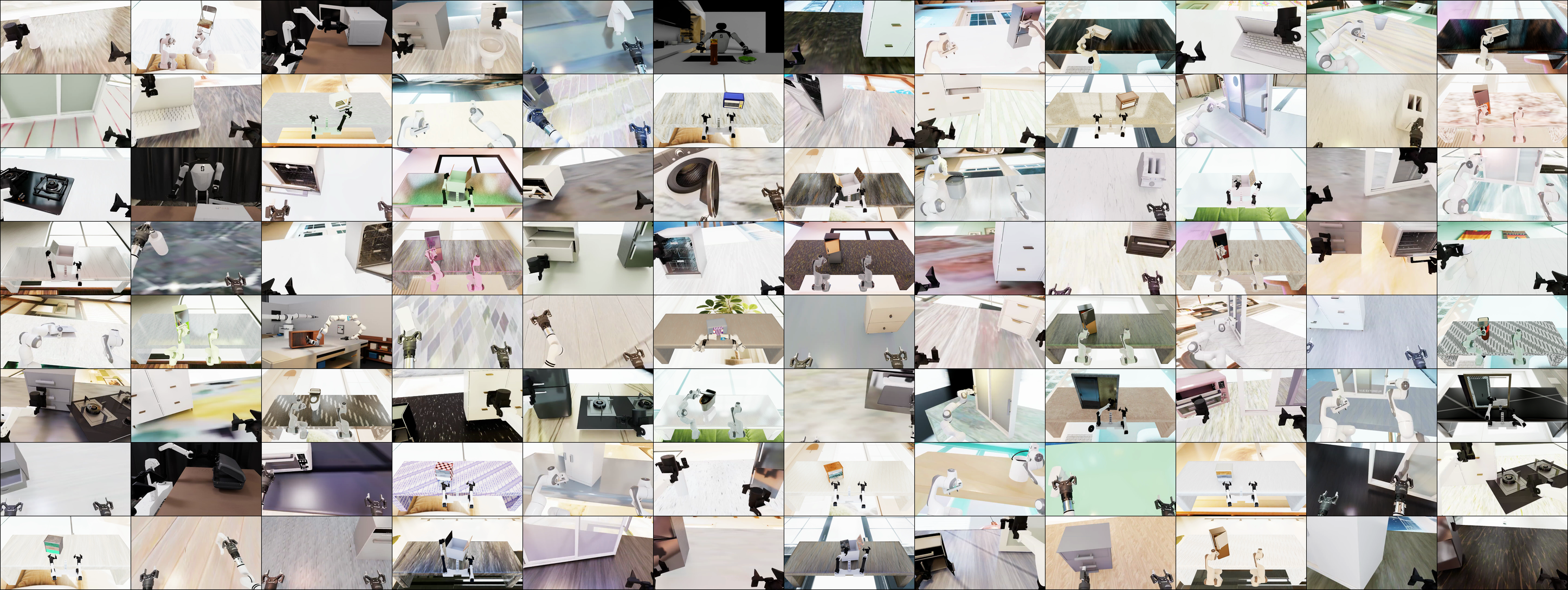}
\caption{Representative samples from \ourdataset across diverse scenes, objects, and manipulation tasks.}
\label{fig:app:dataset_illustration}
\end{figure}

We show representative data samples of \ourdataset in Fig.~\ref{fig:app:dataset_illustration}.

\section{Experiment Details}
\subsection{Robot Platform Details}
\label{app:subsec:robot_platform_details}

We show the details of supported robot platforms in Tab.~\ref{tab:app:robot_platform_details}.
The same task and skill definitions can be reused across these hardware configurations through the setup assembly interface described in Sec.~\ref{subsec:simulation_assets}.

\begin{table}[H]
\centering
\scriptsize
\setlength{\tabcolsep}{5pt}
\renewcommand{\arraystretch}{1.2} 
\renewcommand{\cellgape}{\Gape[2pt]}

\caption{Robot platform details.
Franka is listed as single-arm and dual-arm configurations but is counted as one robot embodiment in the main text.}
\label{tab:app:robot_platform_details}

\begin{tabular}{lcccccccc}
\toprule
\textbf{Platform Name} 
& \textbf{RM75} 
& \textbf{R1Pro} 
& \textbf{AC1} 
& \textbf{\makecell{Franka\\(Single)}} 
& \textbf{\makecell{Franka\\(Dual-Arm)}} 
& \textbf{\makecell{Flexiv\\Rizon 4s}} 
& \textbf{\makecell{Marvin\\M6s}} 
& \textbf{\makecell{Unitree\\H1-2}} \\
\midrule

\textbf{Degrees of Freedom} 
& 16 & 16 & 14 & 8 & 16 & 8 & 16 & 36 \\

\textbf{Camera Number} 
& 3 & 3 & 3 & 2 & 3 & 2 & 3 & 3 \\

\textbf{End-Effector Type} 
& \makecell{Robotiq\\2F-85} 
& \makecell{Self-designed\\Gripper} 
& \makecell{AC1\\Gripper} 
& \makecell{Franka\\Hand} 
& \makecell{Franka\\Hand} 
& \makecell{Robotiq\\2F-85} 
& \makecell{DAS \\ Gripper V3} 
& \makecell{X\\Hand} \\

\textbf{Center Camera Type} 
& \makecell{RealSense\\L515} 
& \makecell{RealSense\\D435} 
& \makecell{RealSense\\D405} 
& \makecell{RealSense\\D435} 
& \makecell{RealSense\\D435} 
& \makecell{RealSense\\D405} 
& \makecell{RealSense\\L515} 
& \makecell{RealSense\\L515} \\

\textbf{Wrist Camera Type} 
& \makecell{RealSense\\D405} 
& \makecell{RealSense\\D405} 
& \makecell{RealSense\\D405} 
& \makecell{RealSense\\D435} 
& \makecell{RealSense\\D435} 
& \makecell{RealSense\\D405} 
& \makecell{DAS\\Fisheye} 
& \makecell{RealSense\\D405} \\

\bottomrule
\end{tabular}
\end{table}

\subsection{Task Details}
\label{app:subsec:task_details}

The real-world evaluation consists of eight task types that expand to twelve evaluation tasks through four cross-platform repetitions.
Bucket handle lifting, toaster-lever pressing, and microwave door closing are evaluated on RM75 and R1Pro, while cabinet door sliding is evaluated on RM75 and AC1.
The remaining task types are evaluated on a single platform.

The manipulation process, initial object state, and success criterion of each task are defined as follows.

\begin{itemize}[leftmargin=*,
nosep
]
    \item \textbf{Bucket handle lifting (RM75 and R1Pro)}: the robot's left arm reaches the compliant handle and lifts the bucket by raising the handle at least $30^\circ$ above the horizontal; the handle starts in its nominal resting configuration, and a trial succeeds when the bucket is lifted by raising the handle to this position.
    \item \textbf{Drawer pulling (AC1)}: the robot's left arm grasps the drawer handle and pulls the drawer out of the cabinet; the drawer starts closed, and a trial succeeds when it is extended by at least $20$\,cm.
    \item \textbf{Dispenser pressing (RM75)}: the robot presses the spring-loaded dispenser button downward; the button starts in its released position, and a trial succeeds when water is dispensed from the container.
    \item \textbf{Cabinet door sliding (RM75 and AC1)}: the robot grasps the round handle on RM75 or the vertical bar handle on AC1 and slides the door to the left; the sliding door starts closed, and a trial succeeds when it is moved to the left by at least $20$\,cm.
    \item \textbf{Toaster lever pressing (RM75 and R1Pro)}: the robot presses the lever from the top all the way down; the lever starts in its upper position, and a trial succeeds when it reaches the bottom without the gripper losing contact during the motion.
    \item \textbf{Microwave door closing (RM75 and R1Pro)}: the robot's left arm pushes the horizontally hinged door shut from an initial opening angle between $60^\circ$ and $80^\circ$, and a trial succeeds when the door is fully closed and the latch is engaged.
    \item \textbf{Microwave knob rotation (AC1)}: the robot's left arm grasps the knob, which points straight up in the initial state, and rotates it clockwise, and a trial succeeds when it has been rotated by at least $70^\circ$.
    \item \textbf{Toolbox lid flipping (AC1)}: the robot's left arm grasps the handle on the upper lid and flips the lid upward to open the box; the upper lid starts closed, and a trial succeeds when its opening angle exceeds $70^\circ$.
\end{itemize}

Across these tasks, the object and robot initial configurations are sampled from small bounded ranges around the nominal reset states, and scene lighting is randomized at each environment reset.

\subsection{Training Details}
\label{app:subsec:training_detail}

We show the hyperparameters used in pretraining and post-training in Tab.~\ref{tab:app:training_hyperparameters}.

\begin{table}[H]
\centering
\setlength{\tabcolsep}{15pt}
\renewcommand{\arraystretch}{1.2}

\caption{Hyperparameters used in pretraining and post-training. For pretraining, we adopt data packing~\citep{zhang2026prts} to form training batches, where the global batch size is estimated according to the average number of samples contained in each data pack.}
\label{tab:app:training_hyperparameters}

\begin{tabular}{lcc}
\toprule
\textbf{Hyperparameters} & \textbf{Pretraining} & \textbf{Post-training (Sim2Real)} \\
\midrule
\textbf{Global Batch Size} & 2048 & 32 \\
\textbf{VLM Learning Rate} & 5e-5 & 1e-5 \\
\textbf{Action Expert Learning Rate} & - & 1e-4 \\
\textbf{Learning Rate Schedule} & Cosine Decay & Cosine Decay \\
\textbf{Minimum Learning Rate} & 5e-6 & 1e-6 \\
\textbf{Warmup Ratio} & 0.01 & 0.01 \\
\textbf{Training Steps} & 278k & 25k \\
\bottomrule
\end{tabular}
\end{table}

\subsection{Evaluation Protocol Details}
\label{app:eval_protocol_details}
We take the AC1 robot as an example to show the definitions of NRMSJ and NMAV.
For AC1, the action vector used for these metrics contains the 12 arm degrees of freedom; the two gripper degrees of freedom are excluded.
For an action chunk, define
\begin{equation}
    A = [\mathbf{a}_0,\mathbf{a}_1,\ldots,\mathbf{a}_{T-1}],
    \qquad
    \mathbf{a}_t \in \mathbb{R}^{D},
    \qquad
    T=50,\ D=12.
\end{equation}

To normalize action dimensions with different scales, we apply task-level
quantile normalization. Let $q_{0.01,d}$ and $q_{0.99,d}$ denote the
1st and 99th percentiles of the $d$-th action dimension, respectively.
The normalized action is defined as
\begin{equation}
    \tilde{a}_{t,d}
    =
    2
    \frac{
        a_{t,d}-q_{0.01,d}
    }{
        q_{0.99,d}-q_{0.01,d}+\varepsilon_{\mathrm{q}}
    }
    -1,
    \qquad
    \varepsilon_{\mathrm{q}}=10^{-8}.
\end{equation}

All metrics below are computed using the quantile-normalized actions
$\tilde{\mathbf{a}}_t$.

\textbf{Normalized RMS Joint Jerk (NRMSJ).}
The third-order finite difference of the normalized action trajectory is
defined as
\begin{equation}
    j_{t,d}
    =
    \Delta^3 \tilde{a}_{t,d}
    =
    \tilde{a}_{t+3,d}
    -3\tilde{a}_{t+2,d}
    +3\tilde{a}_{t+1,d}
    -\tilde{a}_{t,d},
    \qquad
    t=0,\ldots,T-4,
    \quad
    d=1,\ldots,D .
\end{equation}

The normalized RMS joint jerk is computed as
\begin{equation}
    \operatorname{NRMSJ}(A)
    =
    \sqrt{
        \frac{1}{(T-3)D}
        \sum_{t=0}^{T-4}
        \sum_{d=1}^{D}
        j_{t,d}^{2}
    }.
\end{equation}

\textbf{Normalized Mean Action Variation (NMAV).}
Let
\begin{equation}
    \tilde{\mathbf{a}}_t
    =
    [
        \tilde{a}_{t,1},
        \ldots,
        \tilde{a}_{t,D}
    ]^{\top}
\end{equation}
denote the quantile-normalized action vector at time step $t$.
The normalized mean action variation is defined as the mean L2 distance
between consecutive quantile-normalized action vectors:
\begin{equation}
    \operatorname{NMAV}(A)
    =
    \frac{1}{T-1}
    \sum_{t=0}^{T-2}
    \left\|
        \tilde{\mathbf{a}}_{t+1}
        -
        \tilde{\mathbf{a}}_{t}
    \right\|_{2}.
\end{equation}

\textbf{Real-world evaluation protocol.}
Fig.~\ref{fig:real_world_tasks} shows the twelve real-world tasks, and Tab.~\ref{tab:app:robot_platform_details} lists the hardware configuration of the three platforms.
For the sim-to-real experiment, each policy is post-trained on the full set of 500 task-specific synthetic demonstrations.
For the scaling experiment, nested subsets of 500, 200, 100, 50, 10, and 1 demonstrations are sampled recursively from the full dataset.
Each policy is then evaluated directly on the real robot.
Success rate is computed over 15 independent trials per policy and task.
Action chunk quality is additionally measured using the NRMSJ and NMAV metrics defined above.